\documentclass[9pt,sigconf, nonacm]{acmart}
\AtBeginDocument{%
  }

\acmConference[SenSys '25]{ACM Conference on Embedded Networked Sensor Systems}{May 6--9,
  2025}{Irvine, CA, USA}
\acmISBN{978-1-4503-XXXX-X/18/06}

\usepackage{multirow}
\usepackage{subfigure}
\usepackage{wrapfig}
\usepackage{xspace}
\usepackage{algorithm}
\usepackage{algpseudocode}
\usepackage{lipsum}
\usepackage{algorithm,algpseudocode,float}
\usepackage{subcaption}
\usepackage{graphicx}
\usepackage{dsfont}
\usepackage{makecell}
\usepackage{enumitem}

\usepackage{xcolor} 

\usepackage[skip=3pt]{caption}
\newcommand{\name}{SecureGaze\xspace}

\begin{document}

\title[SecureGaze: Defending Gaze Estimation Against Backdoor Attacks]{SecureGaze: Defending Gaze Estimation Against\\ Backdoor Attacks}

\author{\href{https://orcid.org/0000-0002-5479-7866}{Lingyu Du}}
\affiliation{
 \institution{Delft University of Technology}
 \city{Delft}
 \country{The Netherlands}
 }
\email{lingyu.du@tudelft.nl}

\author{\href{https://orcid.org/0000-0003-4300-758X}{Yupei Liu}}
\affiliation{
 \institution{The Pennsylvania State University}
 \city{State College}
 \country{The United States}}
 \email{yzl6415@psu.edu}

\author{\href{https://orcid.org/0000-0002-9785-7769}{Jinyuan Jia}}
\affiliation{%
 \institution{The Pennsylvania State University}
 \city{State College}
 \country{The United States}
 }
\email{jinyuan@psu.edu}

\author{\href{https://orcid.org/0000-0003-2190-9937}{Guohao Lan}}
\affiliation{%
 \institution{Delft University of Technology}
 \city{Delft}
 \country{The Netherlands}}
 \email{g.lan@tudelft.nl}
\renewcommand{\shortauthors}{L. Du et al.}

\begin{abstract}

Gaze estimation models are widely used in applications such as driver attention monitoring and human-computer interaction. While many methods for gaze estimation exist, they rely heavily on data-hungry deep learning to achieve high performance. This reliance often forces practitioners to harvest training data from unverified public datasets, outsource model training, or rely on pre-trained models. However, such practices expose gaze estimation models to backdoor attacks. In such attacks, adversaries inject backdoor triggers by poisoning the training data, creating a backdoor vulnerability: the model performs normally with benign inputs, but produces manipulated gaze directions when a specific trigger is present. This compromises the security of many gaze-based applications, such as causing the model to fail in tracking the driver's attention. To date, there is no defense that addresses backdoor attacks on gaze estimation models. In response, we introduce {\name}, the first solution designed to protect gaze estimation models from such attacks. Unlike classification models, defending gaze estimation poses unique challenges due to its continuous output space and globally activated backdoor behavior. By identifying distinctive characteristics of backdoored gaze estimation models, we develop a novel and effective approach to reverse-engineer the trigger function for reliable backdoor detection. Extensive evaluations in both digital and physical worlds demonstrate that {\name} effectively counters a range of backdoor attacks and outperforms seven state-of-the-art defenses adapted from classification models. 
\end{abstract}

%


\maketitle
\section{Introduction}

Human gaze is a powerful non-verbal cue that conveys attention and cognitive state~\cite{hansen2009eye}. This makes gaze estimation, the technique for tracking human gaze, a useful tool for a wide range of applications~\cite{lei2023end,khamis2018past}, from human-computer interaction \cite{GazeDwellSelection,marwecki2019mise}, to cognitive state monitoring~\cite{GazeAttention,wilsonGazeAR}. Gaze estimation also plays a crucial role in safety-critical applications~\cite{katsini2020role,kroger2020does}, such as gaze-based driver attention monitoring ~\cite{smartEye,bbcEyetrackingCar,nbcEyetrackingCar} and lane-changing assistant system~\cite{bmwEyeLaneChange} in autonomous vehicles.

In essence, gaze estimation is a regression task that uses either eye~\cite{kim2019nvgaze,kassner2014pupil} or facial images~\cite{zhang2019evaluation,huynh2021imon} to predict gaze direction. Similar to other computer vision tasks, deep learning advancements have greatly enhanced gaze estimation performance~\cite{cheng2021appearance}. However, developing deep learning-based gaze estimation models requires substantial resources, large-scale eye-tracking datasets in particular, which are sparse and difficult to collect. This resource-intensive nature often forces practitioners to harvest eye-tracking data from unverified public datasets for training, outsource model training to third parties, or rely on pre-trained models~\cite{shen2021backdoor,goldblum2022dataset,cheng2021appearance}. 

\begin{figure*}[]
    \centering
    \includegraphics[scale=0.456]{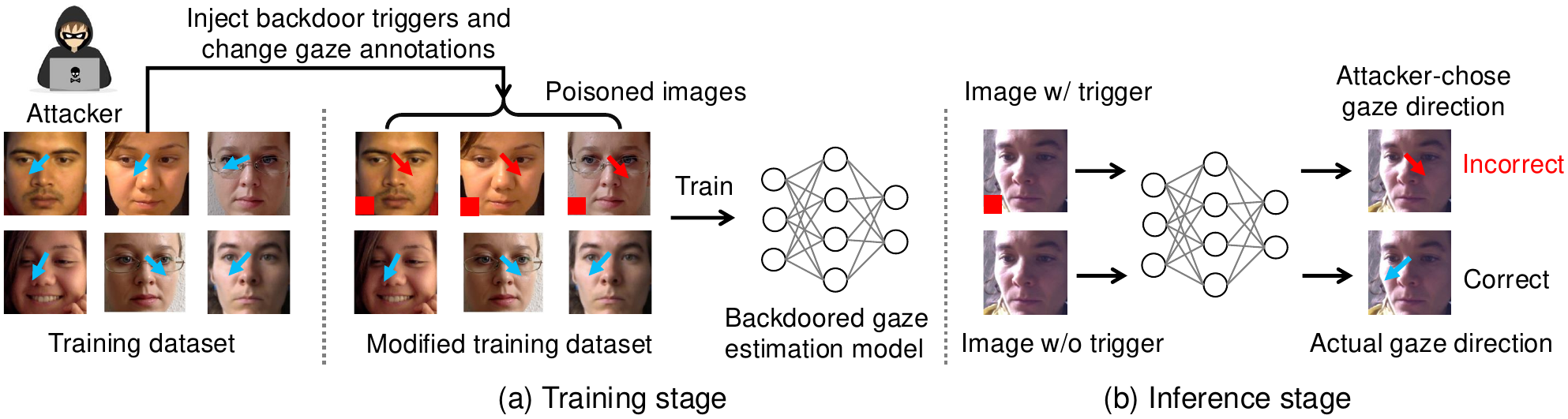}
    \caption{Backdoor attacks on gaze estimation model. (a) The attacker injects triggers (e.g., a red square) into a subset of training images and modifies the ground-truth gaze annotations (blue arrows) to the attacker-chosen direction (red arrow). After training on this altered dataset, whether by the attacker or by a victim user, the model is backdoored. (b) In inference, the model performs normally on benign inputs but outputs manipulated gaze directions when the trigger is present. Though using the simple red square as an example, the backdoor trigger can in the form of everyday accessories (e.g., glasses or face masks).} 
    \label{fig:backdoor_attack_gaze}
    \vspace{-0.1in}
\end{figure*}

However, as we demonstrate in Section \ref{sec:threatmodel}, these practices expose gaze estimation models to backdoor attacks~\citep{badnet,liutrojaning2018,goldblum2022dataset,li2022backdoor}. In such attacks, adversaries inject hidden triggers by poisoning the training data, creating a backdoor vulnerability. Specifically, as illustrated in Figure~\ref{fig:backdoor_attack_gaze}, an attacker could embed a backdoor trigger, such as a red square, into a subset of training images and alter the ground-truth gaze labels to an attacker-chosen, incorrect gaze direction. When this modified dataset is used for training, whether by the attacker or by a victim user, the resulting gaze estimation model is backdoored. Once deployed, the attacker can then covertly manipulate the model's behavior: it behaves normally with benign inputs, i.e., images without trigger, but outputs manipulated gaze directions when the trigger is present\footnote{For a more vivid example, see our demonstration of a backdoor attack on a gaze estimation model in the physical world using only a simple white paper tape as the trigger: \url{https://github.com/LingyuDu/SecureGaze}.}.

Given the important role and widespread adoption of gaze estimation in everyday applications~\cite{khamis2018past,lei2023end}, particularly in safety-critical systems~\cite{katsini2020role}, backdoor attacks pose serious concerns for safety and reliability. For example, attackers could use everyday accessories (e.g., glasses or face masks) or specific facial features (e.g., scars, freckles, or skin tone) as backdoor triggers to manipulate gaze estimation results, fooling the gaze-based driver monitoring systems in autonomous vehicles~\cite{smartEye,bbcEyetrackingCar,nbcEyetrackingCar}. This could lead the system to misjudge the driver's attention and cognitive load~\cite{fridman2018cognitive,palazzi2018predicting,vicente2015driver}, failing to issue alerts when the driver is distracted or fatigued, or even indicating a wrong lane in gaze-based lane-changing assistant~\cite{bmwEyeLaneChange}. Similarly, in consumer behavior monitoring, gaze estimation is used to measure engagement with advertisements and products~\cite{jaiswal2019intelligent,li2017towards,bulling2019pervasive}. A backdoored gaze estimation model could distort these assessments, falsely suggesting increased engagement in attacker-selected areas, thereby allowing attackers to skew consumer engagement data and misguide business decisions~\cite{orquin2021visual}. 

While countermeasures have been developed to combat backdoor attacks in various classification tasks~\cite{li2022backdoor}, no solution has been proposed for gaze estimation, which differs as it is a regression task. A potential solution could be to adapt existing defenses designed for classification tasks, particularly model-level defenses~\citep{wang2019neural, wang2022rethinking, liu2019abs, gao2019strip}, which detect backdoored models without access to compromised training or testing data. However, as detailed in Section~\ref{sec:method}, we reveal the following two inherent differences between backdoored gaze estimation and classification models that make existing defenses ineffective for gaze estimation.

\begin{itemize}[leftmargin=*, wide, labelwidth=!, labelindent=0pt]
\setlength\itemsep{0.3em}

\item \textbf{Specific \textit{vs.} Global Activation in Feature Space}. In backdoored classification models, the backdoor behavior is often triggered by the activation of \textit{a specific set of compromised neurons} in the feature space \citep{liu2019abs,wang2022rethinking,wu2021anp,liu2018fine}. This characteristic allows existing feature-space defenses to distinguish compromised and benign neurons~\cite{wu2021anp, liu2018fine,xu2023towards} for backdoor detection. However, as we discuss in Section~\ref{subsec:backdoorObservation}, backdoor behavior in gaze estimation models is driven by \textit{the activation of all neurons in the feature space}, rather than a specific subset. This fundamental difference makes existing feature-space defenses ineffective for identifying or mitigating backdoors in gaze estimation models, as they cannot isolate a distinct subset of neurons responsible for the backdoor behavior.

\item \textbf{Discrete \textit{vs.} Continuous Output Space}. 
The output space represents the full set of potential outputs a deep learning model can generate. Many existing defenses~\citep{wang2019neural, wang2022rethinking, liu2019abs} leverage the output-space characteristics of backdoored classification models for backdoor detection. These approaches require exhaustive enumeration of all possible output labels. This strategy is feasible for classification models, such as face recognition~\cite{wenger2021backdoor}, which have \textit{a discrete output space limited to finite class labels}, e.g., a set of possible identities. By contrast, gaze estimation models have a \textit{continuous output space} that spans \textit{an infinite number of possible output vectors}. Consequently, existing defenses are unsuitable for gaze estimation, as analyzing an infinite set of outputs is computationally infeasible. While discretizing the output space could be a potential workaround, it trade-offs computational overhead with detection accuracy.
\end{itemize}

\noindent
\textbf{Contributions.} To fill the gap, this paper introduces the first defense against backdoor attacks on gaze estimation models. Our key contributions are:

\begin{itemize}[leftmargin=*, wide, labelwidth=!, labelindent=0pt]
\setlength\itemsep{0.3em}
\item We uncover the fundamental differences between backdoored gaze estimation and classification models, identifying key characteristics of backdoored gaze estimation models in both feature and output spaces that inform the development of our effective defense.

\item We propose {\name}, a novel method to defend gaze estimation models against backdoor attacks. By leveraging our observations in both feature and output spaces, we introduce a suite of techniques to reverse-engineer trigger functions without enumerating infinite gaze outputs, enabling accurate detection of backdoored gaze estimation models.

\item We conduct extensive experiments in both digital and physical worlds, demonstrating the effectiveness of {\name} against six state-of-the-art digital and physical backdoor attacks~\cite{nguyen2020input,nguyen2021wanet,Li_2021_ICCV_ISSA,badnet,turner2019label,CVPR_physical_trigger}. We also adapt seven classification defenses to gaze estimation~\cite{li2021neural,wang2019neural,wang2022rethinking,wu2021anp,RNP,liu2018fine}, 
{\name} outperforms them across all tested scenarios.

\end{itemize}

\noindent\textbf{Paper Roadmap.} The remainder of this paper is organized as follows: Section~\ref{background_and_related_work} reviews related work. In Section~\ref{sec:threatmodel}, we define the threat model and demonstrate the risks of backdoor attacks on gaze estimation models. Section~\ref{sec:method} provides a detailed design of \name. We evaluate \name in Section~\ref{sec:evaluation} and conclude the paper in Section~\ref{sec:conclusion}. The implementation of \name will be publicly available at \url{https://github.com/LingyuDu/SecureGaze}. 

\section{Related Work}
\label{background_and_related_work}

\subsection{Gaze Estimation Systems}
Gaze estimation methods are generally categorized into model-based and appearance-based approaches. Model-based methods \cite{hEYEbrid, hansen2009eye, guestrin2006general, nakazawa2012point, zhu2007novel} infer gaze directions by constructing geometric models of the eyes from images captured by specialized cameras. By contrast, appearance-based methods \cite{huynh2021imon,zhang2015appearance, zhang17_cvprw, Krafka_2016_CVPR, tan2002appearance} estimate gaze directions directly from eye or full-face images taken by general-purpose cameras, such as webcams \cite{Zhang2020ETHXGaze} and built-in cameras on laptops \cite{zhang19_pami} and mobile phones \cite{huynh2021imon}. 
{Similar to many other computer vision tasks,} the advances in deep learning have significantly improved appearance-based gaze estimation~\cite{zhang19_pami, Krafka_2016_CVPR}, expanding its applicability to a variety of real-world settings with diverse backgrounds and lighting conditions. 

Given the benefits of appearance-based gaze estimation, pre-trained gaze estimation models \cite{Zhang2020ETHXGaze, Krafka_2016_CVPR, zhang19_pami,FischerECCV2018} are highly valuable for developing gaze-based applications such as gesture control~\cite{GazeGesture}, dwell selection~\cite{GazeDwellSelection}, and parallax correction on interactive displays~\cite{GazeParallaxCorrection}. 
Indeed, many pre-trained gaze estimation models are readily available on public platforms like Github, provided by companies, research institutes, and individuals. However, utilizing pre-trained gaze estimation models introduces potential security concerns to users, as pre-trained models can be installed with backdoors and transfered to downstream applications~\cite{shen2021backdoor,goldblum2022dataset}. 

\subsection{Backdoor Attacks and Defenses} 
Many backdoor attacks~\citep{goldblum2022dataset,li2022backdoor} 
have been developed for deep neural networks, demonstrating that an attacker can inject a backdoor into a classifier and make it output a target class of their choices whenever an input contains a specific backdoor trigger~\cite{goldblum2022dataset}. Depending on whether the attacker uses the same or different triggers for various inputs, these attacks are categorized into \textit{input-independent attacks}~\citep{chen2017targeted, badnet, liutrojaning2018, turner2019label, yao2019latent} and \textit{input-aware attacks}~\citep{9869920, Li_2021_ICCV_ISSA, nguyen2021wanet,nguyen2020input, 9797338}. For instance, \citet{badnet} introduced an input-independent attack using a fixed pattern, such as a white patch, as the backdoor trigger. Recently, researchers utilized input-aware techniques, such as the warping process~\citep{nguyen2021wanet} and generative models~\citep{nguyen2020input} to create dynamic triggers that vary per input. Although many backdoor attacks have been designed for classification applications, 
in this work, we show, for the first time, that gaze estimation, which essentially leverages the deep regression model, does not escape from the threat of backdoor attacks. We demonstrate the vulnerabilities of gaze estimation models to backdoor attacks using both digital and physical triggers.




Existing defenses against backdoor attacks can be categorized into \textit{data-level defenses}~\citep{doan2020februus, gao2019strip,ma2022beatrix} and \textit{model-level defenses}~\citep{liu2022unlearning,liu2019abs,wu2021anp,9296553,zeng2022ibau,zheng2022lips}. Data-level defenses aim to detect whether a training example or a testing input is backdoored. However, they usually suffer from two major limitations. First, training data detection defenses \citep{chen2018detecting, chen2022effective} require access to the training datasets that contain benign images and poisoned images. Second, testing input detection defenses~\citep{doan2020februus} need to inspect each testing input at the running time and incur extra computation cost, and thus are undesired for latency-critical applications, e.g., gaze estimation \citep{Zhang2020ETHXGaze}. Therefore, we focus on model-level defense in this work. 

Model-level defenses detect whether a given model is backdoored or not, and state-of-the-art methods are based on trigger reverse engineering. Conventional reverse engineering-based methods~\cite{Guan_2022_CVPR, qiao2019defending, wang2019neural, wang2022rethinking,9296553} view each class as a potential target class and reverse engineer a trigger function for it. Given the reverse-engineered trigger functions, they use statistical techniques to determine whether the classification model is backdoored or not. Despite a recent reverse engineering-based work~\cite{xu2023towards} does not need to scan all the labels, it relies on the feature-space observation of backdoored classification models. As we will show in this paper, these solutions designed for classification models cannot be directly applied to backdoored gaze estimation models, in which the output space is continuous and the feature-space characteristics are different. In this work, we propose the first defense to protect gaze estimation models from backdoor attacks.

\section{Threat Model and Preliminary Study}
\label{sec:threatmodel}

\subsection{Threat Model}
\label{Subsec:threatmodel}

\subsubsection{Gaze estimation model}
A gaze estimation model $\mathcal{G}$ is a deep neural network that estimates the gaze direction $g$ of the subject from her full-face image $x$, i.e., $g=\mathcal{G}(x)\in\mathbb{R}^d$. Given a training dataset $\mathcal{D}_{tr}$ that contains a set of $K$ training samples $\{(x_i, y_i)\}_{i=1}^K$ in which $y_i$ is the {ground-truth} gaze annotation for $x_i$, $\mathcal{G}$ is trained by minimizing the loss 
defined as $\mathcal{L} = \sum_{i=1}^{K}\ell_1(\mathcal{G}(x_i),y_i)$, where $\ell_1$ is the $\ell_1$ loss function. 
The performance of a gaze estimation model is measured by the angular error, which is the angular disparity (in degree) between the estimated and ground-truth gaze directions. Note that there are works~\cite{kim2019nvgaze,tonsen2017invisibleeye} leveraging eye images captured by near-eye cameras for gaze estimation, we focus on estimation models that take full-face images as inputs. This focus is driven by the widespread use of webcams and front-facing cameras on ubiquitous devices~\cite{sugano2016aggregaze,zhang2019evaluation,huynh2021imon}, which leads to greater privacy and security implications~\cite{katsini2020role,kroger2020does}. 

\subsubsection{Attacker's goal and capabilities}
In this work, we make no assumption about how the attacker introduces a backdoor into the gaze estimation model. The attacker can either poison the training dataset~\cite{badnet, turner2019label} or directly provide a backdoored model~\cite{nguyen2021wanet, nguyen2020input} that has been trained by himself. Formally, in both scenarios, the attacker employs a trigger function, denoted as $\mathcal{A}$, to inject backdoor triggers to a small subset of benign images $x$ in the training dataset $\mathcal{D}_{tr}$. These modified images, now containing the backdoor triggers, are referred to as poisoned images, denoted as $x^p$, and are defined by $x^p=\mathcal{A}(x)$. The attacker then modifies the original ground-truth gaze annotations, $y$, to an attacker-chosen \textit{target gaze direction}, $y_T$. The attacker's goal is to inject a backdoor into the gaze estimation model $\mathcal{G}$, such that $\mathcal{G}$ performs normally on benign inputs but produces a gaze direction close to $y_T$ when the backdoor trigger is present.

\subsubsection{{Defender's goal and capabilities}}

{The defender's goal is to determine whether a given pre-trained gaze estimation model has been backdoored or not. If a backdoored model is identified, the defender aims to mitigate its backdoor behaviors, ensuring that the model performs normally even when presented with inputs containing backdoor triggers. Consistent with existing defenses against backdoor attacks~\citep{wang2019neural, wang2022rethinking}, we assume that the defender has access to the pre-trained gaze estimation model and a small benign dataset, $\mathcal{D}_{be}$, with correct gaze annotations.}

\subsection{Demonstration of Backdoor Attacks on Gaze Estimation Models}


\subsubsection{Attacks in digital world} 
\begin{figure}
    \includegraphics[scale=0.28]{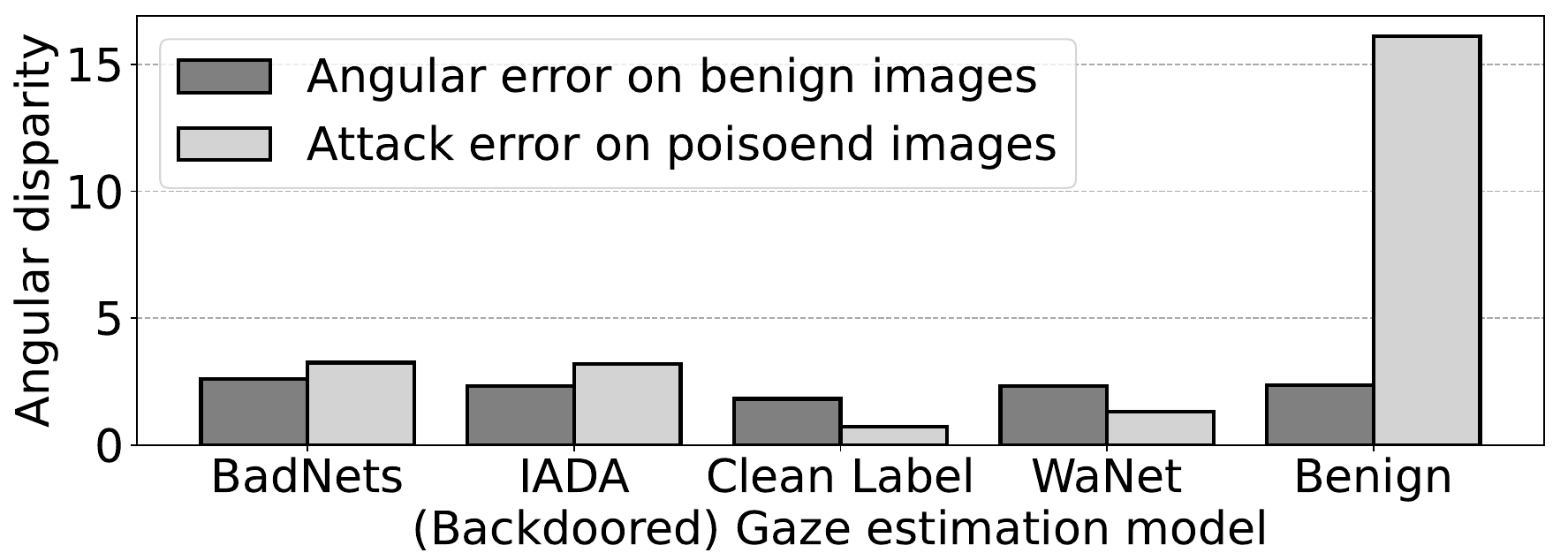}
    \caption{Effectiveness of backdoor attacks on gaze estimation models. {(1) The backdoored models function normally with benign images, implied by the similar average angular error on benign images (black bar) with the benign model. (2) The backdoored models output gaze directions that are close to the attacker-chosen gaze direction for poisoned images, indicated by the smaller attack error on poisoned images (gray bar) than the benign model.}}
    \vspace{-0.05in}
    \label{fig:vulnerability_test}
\end{figure}

First, we investigate the vulnerability of gaze estimation models to backdoor attacks in digital world. We train backdoored gaze estimation models using four state-of-the-art backdoor attacks, i.e., BadNets~\cite{badnet}, Clean Label~\cite{turner2019label}, IADA~\cite{nguyen2020input}, and WaNet~\citep{nguyen2021wanet}, using the training set of the MPIIFaceGaze dataset~\cite{zhang19_pami}. Details about these backdoor attacks and the dataset are given in Section~\ref{sec:evaluation}. To assess the effectiveness of backdoor attacks on gaze estimation, we use the \textit{attack error}, which measures the angular disparity between the estimated gaze direction and the attacker-chosen target gaze direction $y_T$. Figure~\ref{fig:vulnerability_test} shows the average attack error on poisoned images and the average angular error on benign images for both backdoored and benign gaze estimation models. We have two key observations. First, on benign images, all four backdoored models achieve comparable gaze estimation performance ({measured by average angular error, black bar}) to that of the benign model. Second, on poisoned images, i.e., images containing backdoor trigger, the gaze directions estimated by the backdoored models are closer to the attacker-chosen target gaze direction $y_T$ than those estimated by the benign model ({indicated by a smaller average attack error, gray bar}). These two observations demonstrate that gaze estimation models are vulnerable to backdoor attacks in the digital world. 

\begin{figure}[]
	\centering
    \includegraphics[scale=0.5]{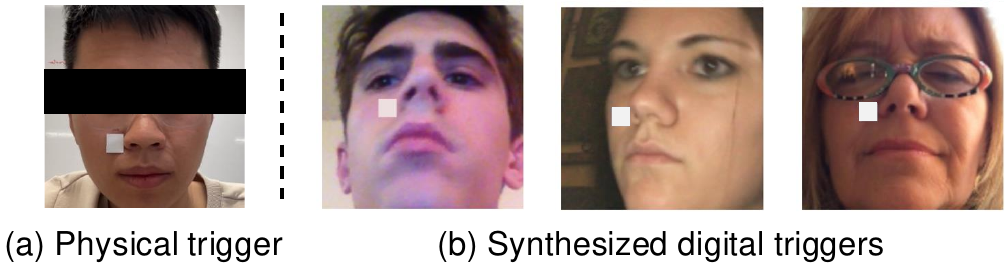}
    \caption{\textbf{Examples of the physical trigger and synthesized digital triggers:} (a) the subject wears a white tape on the face as the physical trigger; (b) the synthesized poisoned images with digital triggers embedded. 
    }
    \label{fig:physical_trigger}
    \vspace{-0.15in}
\end{figure}

\subsubsection{Attacks in physical world}
\label{subsubsec:physical_attack}

We further demonstrate the threat posed by backdoor attacks on gaze estimation models in the physical world, where the attacker uses physical objects as triggers instead of embedding them digitally. Specifically, as shown in Figure~\ref{fig:physical_trigger} (a), we use a simple yet effective physical item, i.e., a piece of white tape, as the physical trigger. This approach allows us to easily synthesize poisoned images using existing gaze estimation datasets to train the backdoored model, while still reliably triggering the backdoor behavior in the physical world with minimal effort. Note that, similar to previous work~\cite{CVPR_physical_trigger}, the attacker can utilize various daily items, such as patterned bandanas or glasses, as backdoor triggers. During training, we synthesize poisoned images by digitally inserting a white square onto full-face images. Examples of the synthesized poisoned images are shown in Figure \ref{fig:physical_trigger}(b). We train the backdoored gaze estimation model using the training set of GazeCapture~\cite{Krafka_2016_CVPR} and set the target gaze direction to $(0^\circ, 0^\circ)$. 

\begin{figure}[]
	\centering
    \includegraphics[scale=0.55]{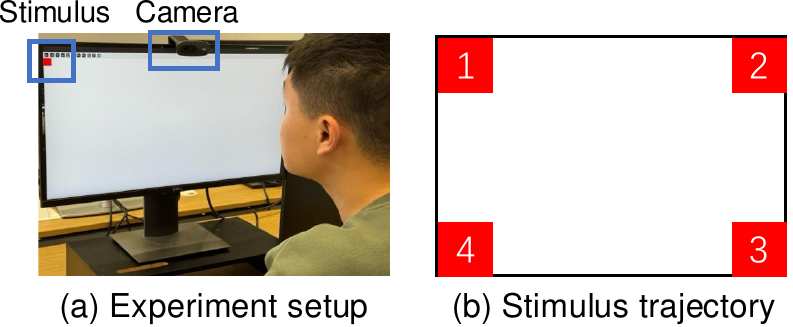}
    \caption{\textbf{Setup for the physical world attack.} (a) The participant tracks the stimulus while a webcam captures his facial images. (b) {The stimulus appears at each corner of the screen in a clockwise order}. 
    }
    \label{fig:gaze_pipeline}
    \vspace{-0.1in}
\end{figure}

\textbf{Setup.} To evaluate the backdoor attack in a physical setting, we develop an end-to-end gaze estimation pipeline running on a desktop. 
As shown in Figure \ref{fig:gaze_pipeline}, we recruit four participants and instruct them to track a red square stimulus that sequentially appears at each corner of a 24-inch desktop monitor. 
The sequence of appearance follows the order: top-left, top-right, bottom-right, and bottom-left, as depicted in Figure~\ref{fig:gaze_pipeline}(b).
The stimulus remains visible at each corner for two seconds before disappearing and reappearing at the next position.
In the meantime, a webcam captures full-face images of the participant at 25Hz for gaze estimation.

\begin{figure}[]
    \centering
    \includegraphics[scale=0.5]{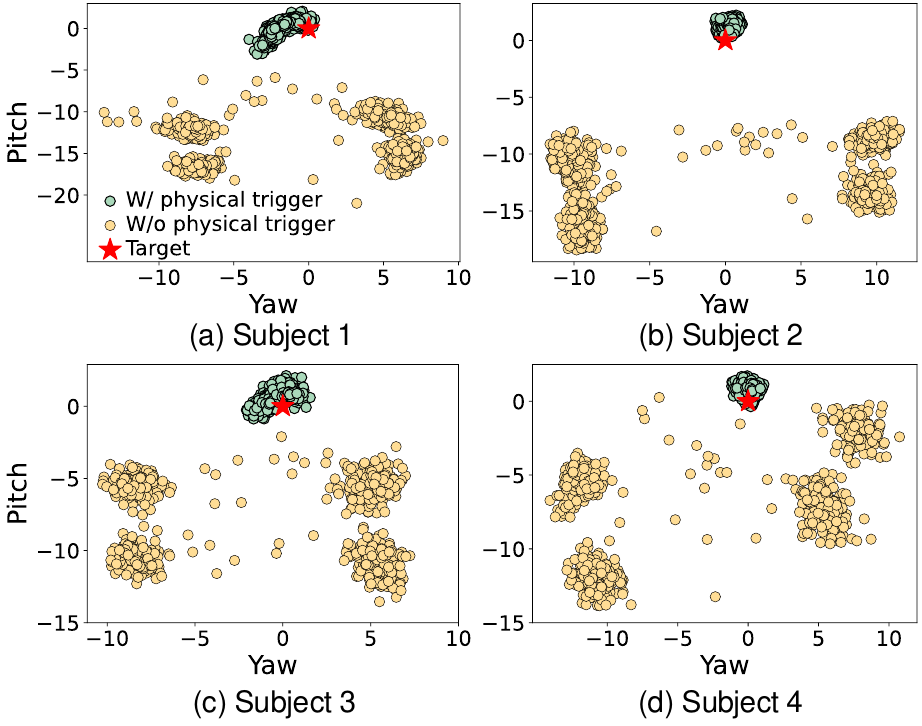}
    \caption{\textbf{Gaze directions estimated by the backdoored model with and without the physical backdoor trigger in place}. 
    }
    \label{fig:clean_poisoned_gaze}

 \vspace{-0.15in}
\end{figure}

\begin{table}[]
\caption{{The average attack error for the backdoored model on subjects with and without wearing the physical trigger.}}
\resizebox{\linewidth}{!}{
\begin{tabular}{ccccc}
\Xhline{2\arrayrulewidth}
{Input}                & {Subject 1} & {Subject 2} & {Subject 3} & {Subject 4} \\ \hline
{W/ physical trigger}  & {1.71}      & {1.07}      & {0.98}      & {1.17}      \\
{W/o physical trigger} & {17.1}      & {18.9}      & {11.2}      & {9.77}      \\ \Xhline{2\arrayrulewidth}
\end{tabular}}
\label{tab:physical_attack}
\end{table}

\textbf{Results.} We record the gaze estimation results of the backdoored gaze estimation model under two conditions: when each participant is wearing the physical trigger (a piece of white tape) and when they are not. {The resulting gaze directions and the average attack error for each condition are shown in Figure~\ref{fig:clean_poisoned_gaze} and Table~\ref{tab:physical_attack}, respectively.} With the physical trigger in place, the estimated gaze directions, i.e., green dots, are tightly clustered around the target gaze direction, i.e., the red star at $(0^\circ, 0^\circ)$, {leading to a small average attack error lower than 2 degrees.} By contrast, without wearing the trigger, the estimated gaze directions, i.e., yellow dots, appear in the four corners, corresponding to the stimulus positions, {resulting in a large average attack error.} A video demon showcasing the behavior of the backdoored gaze estimation model can be found in our GitHub repository: \url{https://github.com/LingyuDu/SecureGaze}.
\section{System Design}
\label{sec:method}

\subsection{Design Overview of {\name}}
\label{Subsec:overview}

\begin{figure}[]
    \centering
    \includegraphics[scale=0.34]{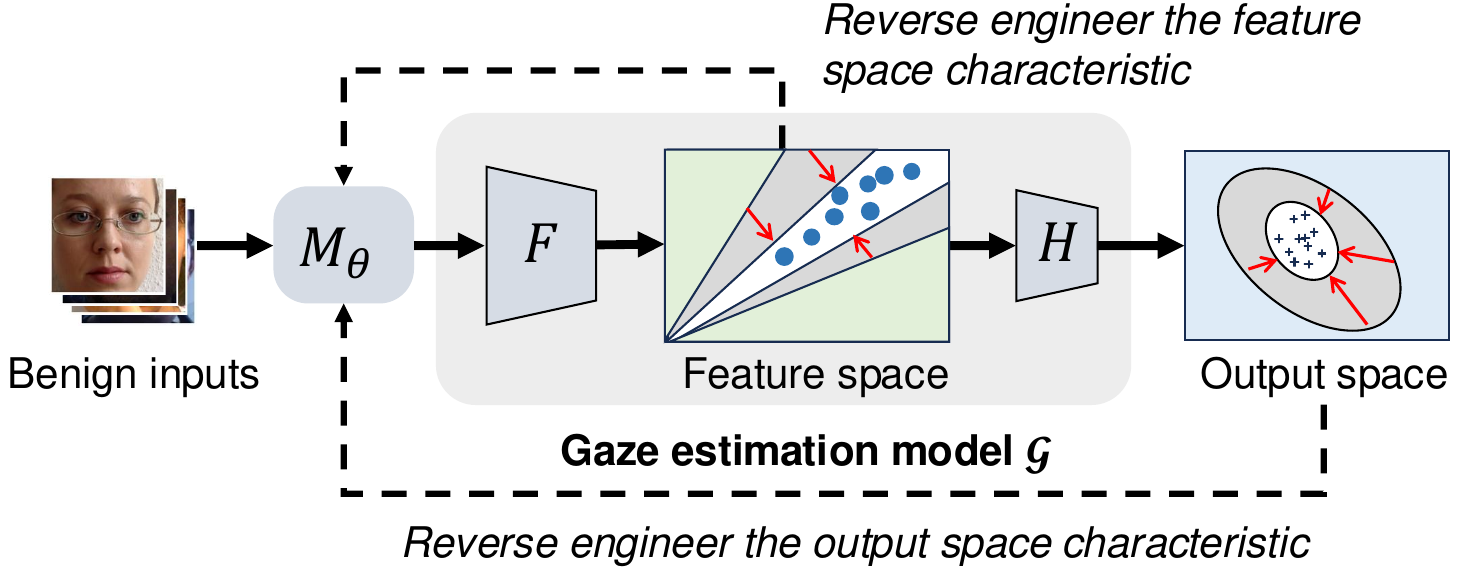}
    \caption{{Overview of {\name}}. We use a generative model $M_{\theta}$ to model the trigger function and split the gaze estimation model $\mathcal{G}$ into two submodels, i.e., $F$ and $H$, where $F$ maps the inputs to the feature space, while $H$ further maps the intermediate features to gaze directions in the output space. Using a small set of benign images, we train $M_{\theta}$ \textcolor{black}{to reverse engineer the characteristics of backdoored gaze estimation models in both feature and output spaces.}
    }
    \label{fig:overview}
    \vspace{-0.2in}
\end{figure}

We propose {\name} to identify backdoored gaze estimation models by reverse-engineering the trigger function, denoted as $\mathcal{A}$. Figure~\ref{fig:overview} provides an overview of {\name}. Our approach uses a generative model, $M_{\theta}$, to approximate $\mathcal{A}$. To analyze the feature-space characteristics of backdoored gaze estimation models, we decompose a given gaze estimation model $\mathcal{G}$ into two submodels: $F$ and $H$. Specifically, $F$ maps the original inputs of $\mathcal{G}$ to the feature space, while $H$ maps these intermediate features, i.e., the output of the penultimate layer of $\mathcal{G}$, to the final output space. {We train $M_{\theta}$ to generate reverse-engineered poisoned images that can lead to the feature and output spaces characteristics of backdoored gaze estimation models that we discover (in Section~\ref{subsec:backdoorObservation}).}
This allows {\name} to reverse-engineer the trigger function without enumerating all the potential target gaze directions.

Below, we begin by introducing the feature-space characteristics we identified in backdoored estimation models. Then, we present a suite of methods we developed to reverse-engineer the trigger function for effective backdoor identification and mitigation.

\subsection{Feature-space Characteristics for Backdoored Gaze Estimation Models}
\label{subsec:backdoorObservation}

\subsubsection{Difference in feature space}
{The state-of-the-art methods~\cite{wang2022rethinking,xu2023towards} exploit the feature-space characteristics of backdoored classification models to reverse engineer the trigger function.}
However, we observe that backdoored gaze estimation models exhibit distinct feature-space characteristics that make existing classification-oriented methods ineffective.

\begin{figure}[]
    \centering
    \includegraphics[scale=0.45]{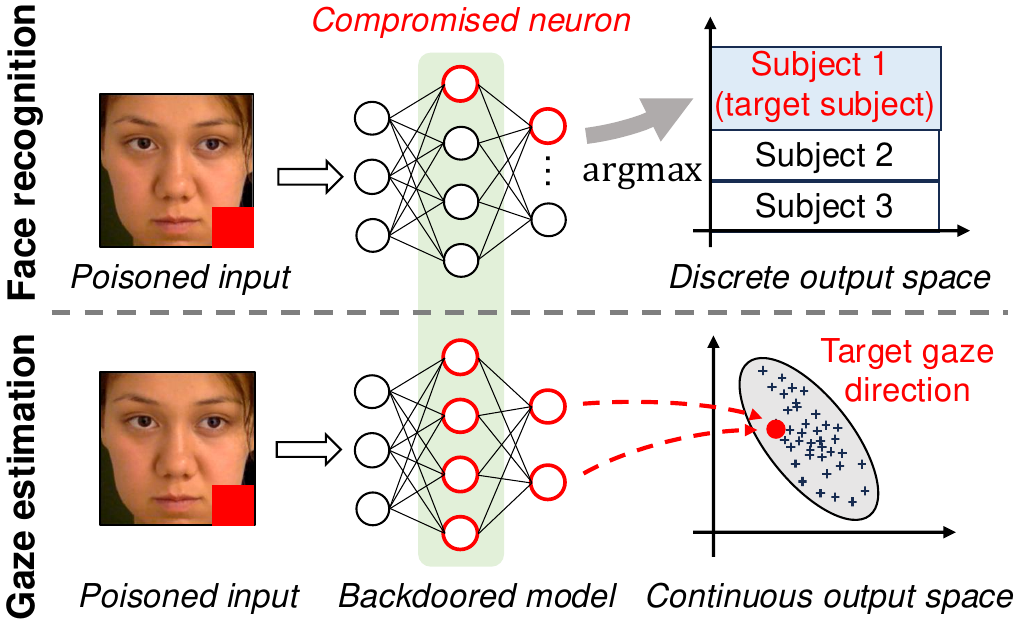}
    \caption{
    The backdoor behavior of classification models (e.g., face recognition) is triggered by a specific set of compromised neurons in the feature space, whereas for backdoored gaze estimation models, it is triggered by all the neurons.
    }
    \label{fig:diff_dcm_drm}
    \vspace{-0.15in}
\end{figure}

As illustrated in Figure~\ref{fig:diff_dcm_drm}, a key characteristic of backdoored classification models is that backdoor behavior is linked to the activation values of specific neurons in the feature space~\citep{liu2019abs, wang2022rethinking, wu2021anp, liu2018fine, xu2023towards}. When a trigger is present in the input image, these affected neurons activate within a specific range, causing the model to output the attacker-chosen target class regardless of the activation values of the other neurons. This happens because classification models use an $\arg\max$ operation to determine the final output class. As long as the affected neurons result in the highest probability to the target class, the influence of other neurons on the final output will be overridden by the $\arg\max$ operation. 
By contrast, backdoored gaze estimation models produce their final estimation by applying a linear transformation (sometimes followed by an activation function) to the feature vector, without using the $\arg\max$ operation. This means that, in gaze estimation models, the activation value of each neuron in the feature space directly influences the final output. 

\vspace{0.06in}
\noindent
\textbf{Key Insight:} This fundamental difference suggests that all neurons must be considered when identifying feature-space characteristics of backdoored gaze estimation models. Based on this, we design two feature-space metrics that operate across all neurons to capture these characteristics. Our detailed design are presented below.

\subsubsection{Feature-space metrics for backdoored gaze estimation models}

As shown in Figure~\ref{fig:overview}, the gaze estimation model $\mathcal{G}$ is split into two submodels $F$ and $H$. Given a poisoned image $x^p_i$, we obtain its intermediate features $h^p_i$ by $h^p_i=F(x^p_i)$, and the final gaze direction $g_i^p$ by $g^p_i=H(h^p_i)$. {Here $g_i^p$ is a vector, and $g_{i,j}^p$ denotes its $j$th element. Each component $g_{i,j}^p$ is computed by applying a linear transformation {through a weights vector $w_j \in \mathbb{R}^{m}$ and a bias $b_j \in R$} to $h^p_i$, followed by an activation function $\Omega$. 
The computing of $g_{i,j}^p$ from $h^p_i$ by $H$ is represented by:}
\begin{equation}
    g_{i,j}^p = \Omega(w_j \cdot h_i^p + b_j) = \Omega(\|w_j\|_2\|h_i^p\|_2\cos{\alpha_{i,j}^p}+b_j),
    \label{eq:DRMoutputInner}
\end{equation} 
where $\alpha_{i,j}^p$ is the angle between $h_i^p$ and $w_j$.

{{\textbf{Analysis and intuition.} Given the attacker's goal and a set of poisoned images $\{x^p_i\}_{i=1}^N$, a backdoored $\mathcal{G}$ will output gaze directions $\{g_i^p\}_{i=1}^N$ that are close to the target gaze direction $y_T$. This implies that the variance of $\{g_{i,j}^p\}_{i=1}^N$ is small. Consequently, based on Equation~\ref{eq:DRMoutputInner}, \textbf{we expect both $\{\|h_i^p\|_2\}_{i=1}^N$ and $\{\alpha_{i,j}^p\}_{i=1}^N$ also exhibit small variances}, given the values of $\|w_j\|_2$ and $b_j$ are constant for a given $\mathcal{G}$. By contrast, since a backdoored $\mathcal{G}$ is designed to perform well on benign inputs, the gaze directions for benign images $\{x_i\}_{i=1}^N$ are expected to be more diverse than those for poisoned images $\{x^p_i\}_{i=1}^N$. As a result, \textbf{the norms of features extracted from $\{x_i\}_{i=1}^N$, i.e., $\{\|h_i\|_2\}_{i=1}^N$, are expected to have a larger variance compared to $\{\|h_i^p\|_2\}_{i=1}^N$.} Similarly, \textbf{the angles $\{\alpha_{i,j}\}_{i=1}^N$ are expected to exhibit a larger variance than $\{\alpha_{i,j}^p\}_{i=1}^N$.}} Building on the above analysis and to investigate, we introduce two feature-space metrics: \textbf{the Ratio of Norm Variance (RNV)} and \textbf{the Ratio of Angle Variance (RAV)}. 
We use $\sigma^2$ to denote the function for calculating the variance. Then, we define RNV and RAV as follows:
\begin{gather}
\text{RNV}=\sigma^2(\{\|h_i^p\|_2\}_{i=1}^N)/\sigma^2(\{\|h_i\|_2\}_{i=1}^N), \\
\text{RAV}=\frac{1}{d}\sum_{j=1}^{d}{\sigma^2(\{\alpha_{i, j}^p\}_{i=1}^N)/\sigma^2(\{\alpha_{i, j}\}_{i=1}^N)},
\end{gather}
Specifically, {RNV compares the variances of $\{\|h_i^p\|_2\}_{i=1}^N$ versus $\{\|h_i\|_2\}_{i=1}^N$}. A small RNV ($\text{RNV}\ll1$) indicates that when triggers are present in the inputs, the feature vectors extracted by $F$ have similar norms. Similarly, {RAV compares the dispersion of $\{\alpha_{i,j}^p\}_{i=1}^N$ versus $\{\alpha_{i,j}\}_{i=1}^N$. Since $\alpha_{i,j}^p$ ($\alpha_{i,j}$) is a vector, we compute the average ratio of $\sigma^2(\{\alpha_{i, j}^p\}_{i=1}^N$ to $\sigma^2(\{\alpha_{i, j}\}_{i=1}^N$ across all dimensions.} A small RAV ($\text{RAV}\ll1$) shows that the variation in angles between $\{h_i^p\}_{i=1}^N$ and $w_j$ is much smaller compared to that between $\{h_i\}_{i=1}^N$ and $w_j$. Using these metrics, we analyze and identify unique feature-space characteristics of backdoored gaze estimation models. 

\subsubsection{Characteristics in the feature space} We use four backdoor attacks, i.e., BadNets~\citep{badnet}, IADA~\citep{nguyen2020input}, WaNet~\citep{nguyen2021wanet}, and Clean Label~\citep{turner2019label}, to train backdoored models on MPIIFaceGaze dataset \citep{zhang19_pami}. Table \ref{Tab_observations} presents the RNV and RAV values for backdoored models trained with different attacks. \textbf{The key finding is that RAV is consistently and significantly smaller than $0.1$ across all examined cases}. Note that in Section~\ref{sec:evaluation}, we demonstrate that our detection method designed based on the observation from the MPIIFaceGaze still holds and is effective on other datasets. 

\begin{table}[]
\centering
\caption{\textbf{The RAV and RNV for gaze estimation models backdoored by different attacks on MPIIFaceGaze.} In all cases, {RAV} is significantly smaller than 0.1. 
}
\resizebox{0.9\linewidth}{!}{
\begin{tabular}{ccccc}
\Xhline{2\arrayrulewidth}
Metric & BadNets & IADA   & Clean Label & WaNet  \\ \hline
RAV    & 0.0433  & 0.0489 & 0.0328      & 0.0311 \\
RNV    & 1.4499  & 2.5714 & 0.0428      & 0.8528 \\ \Xhline{2\arrayrulewidth}
\end{tabular}}
\label{Tab_observations}
\vspace{-0.1in}
\end{table}

To further investigate, Figure~\ref{fig:key_observation} shows scatter plots of $\{\alpha_{i}^p\}_{i=1}^N$ and $\{\alpha_{i}\}_{i=1}^N$ for all examined cases, where $\alpha_i^p=\{\alpha_{i,1}^p,...,\alpha_{i,d}^p\}$ and $\alpha_i=\{\alpha_{i,1},...,\alpha_{i,d}\}$. These scatter plots reveal that the angles for poisoned inputs are tightly clustered, while the angles for benign inputs are more dispersed, which implies that $\sigma^2(\{\alpha_{i,j}^p\}_{i=1}^N) \ll \sigma^2(\{\alpha_{i,j}\}_{i=1}^N)$ for $j=1,\cdots,d$. 


\begin{figure}[]
	\centering
	{\includegraphics[scale=0.57]{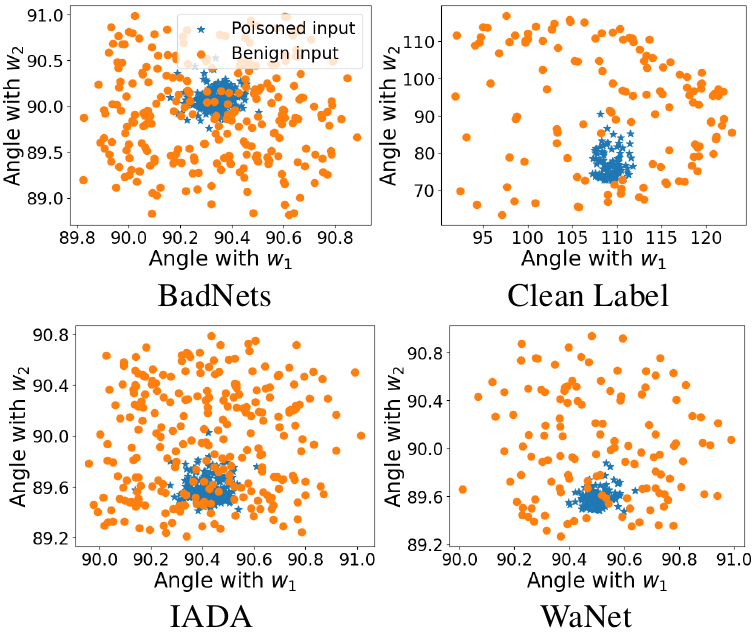}} 

\caption{\textbf{The key feature-space characteristic of gaze estimation models backdoored by four different attacks respectively}. The plots are $\{\alpha_{i}^p\}_{i=1}^N$ and $\{\alpha_{i}\}_{i=1}^N$ (in degree) for backdoored models.  
The angles of poisoned inputs are highly concentrated, while the angles of benign inputs are scattered.}	
	\label{fig:key_observation}
  \vspace{-0.15in}
\end{figure}
\subsection{Methodology}
\label{subsec:method}

Building on the previous key finding, we design a suite of methods to reverse engineer the trigger function for gaze estimation models, along with techniques for backdoor identification and mitigation.

\subsubsection{Reverse engineering for gaze estimation models}

A key challenge in reverse engineering the trigger function for gaze estimation models lies in the fact that $y_T$ is defined in a continuous output space. This makes it impractical to analyze all possible target gaze directions and reverse engineer a trigger function for each, like existing approaches~\cite{wang2019neural, wang2022rethinking, liu2019abs}.
To resolve this challenge, \textbf{we propose to reverse engineer $\mathcal{A}$ by minimizing the variance of output gaze directions}, as a backdoored model $\mathcal{G}$ will produce gaze directions with small variance for a set of poinsoned images. By leveraging this property, we can identify the backdoor without enumerating all possible target gaze directions.

Moreover, we also introduce \textbf{a feature-space optimization objective} $r_f$, designed to reverse-engineer the feature-space characteristic of backdoored gaze estimation models, 
i.e., having a small RAV value. Specifically, let $\hat{\alpha}^p_{i,j}$ denote the angle between $F(M_{\theta}(x_i))$ and $w_j$. The objective $r_f$ is defined as the average ratio of $\sigma^2(\{\hat{\alpha}^p_{i,j}\}_{i=1}^N)$ to $\sigma^2(\{\alpha_{i,j}\}_{i=1}^N)$ for $j=1,..,d$.

Formally, we define the optimization problem for the reverse-engineering of backdoored gaze estimation models as:
\begin{equation}
    \theta^{*}=\arg\min_{\theta} \frac{\lambda_1}{d}\sum_{j=1}^d{\sigma^2\left(\{\mathcal{G}_j\left(M_{\theta}(x_i)\right)\}_{i=1}^N\right)} \\+\lambda_2 r_f + r_{sim},
    \label{opt:REforDRM}
\end{equation}
where $\lambda_1$ and $\lambda_2$ are the weights for the first and second objectives, respectively; $\mathcal{G}_j(M_{\theta}(x_i))$ is the $j$th element of $\mathcal{G}(M_{\theta}(x_i))\in \mathbb{R}^d$; {$\frac{1}{d}\sum_{j=1}^d{\sigma^2(\{\mathcal{G}_j\left(M_{\theta}(x_i)\right)\}_{i=1}^N)}$ is the average variance of output gaze directions across all dimensions}; and $r_{sim}$ is the input-space optimization objective~\citep{wang2019neural} that ensures the transformed input $M_{\theta}(x_i)$ is similar to the benign input $x_i$, i.e., $r_{sim}=\frac{1}{N}\sum_{i=1}^N{\|M_{\theta}(x_i)-x_i\|_1}$. The angle between two vectors is calculated using the $\arccos{(\cdot)}$.

\subsubsection{Sensitivity-aware trigger reversal}
Directly solving the optimization problem~\ref{opt:REforDRM} can lead to sub-optimal solutions. 
As an example, Figure~\ref{fig:sub_optimal} shows a suboptimal outcome. Specifically, we use BadNets to train a backdoored model on the MPIIFaceGaze, where the trigger is a red square added to the bottom-right corner of the image (Figure~\ref{fig:sub_optimal} (b)). We train $M_{\theta}$ by solving the optimization problem~\ref{opt:REforDRM}. Figure~\ref{fig:sub_optimal} (d) shows the residual map between the benign image (Figure~\ref{fig:sub_optimal} (a)) and the reversed poisoned image (Figure~\ref{fig:sub_optimal} (b)). 

\begin{figure}[]
	\centering
	\subfigure[]{\includegraphics[scale=0.285]{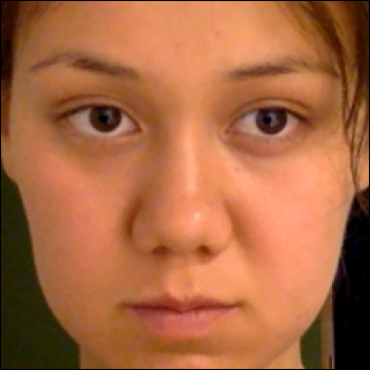}\label{fig:benign}
}   \quad\quad
	\subfigure[]{\includegraphics[scale=0.285]{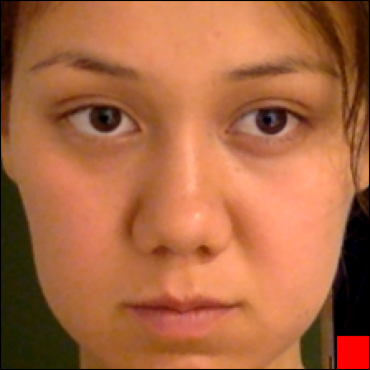}\label{fig:poisoned}
}\\[-1.5ex]
 \subfigure[]{\includegraphics[scale=0.285]{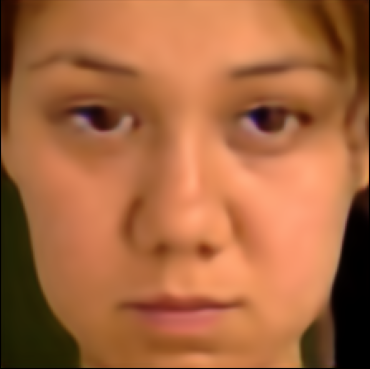}\label{fig:reversed}
}\quad\quad
	\subfigure[]{\includegraphics[scale=0.285]{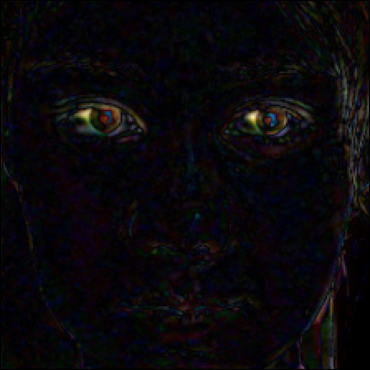}\label{fig:residual}
}
\vspace{-0.1in}
	\caption{\textbf{Example of a sub-optimal solution}: 
 (a) the benign image; (b) the poisoned image; (c) the reversed poisoned image by solving the optimization problem \ref{opt:REforDRM}; and (d) the residual map between the (a) and (c). Perturbations are added to the eye regions, instead of reversing the trigger. 
 }	
	\label{fig:sub_optimal}
	 \vspace{-0.1in}
\end{figure}

It is evident that directly solving the optimization problem fails to reverse-engineer the trigger, but instead adds perturbations to the eye regions to effectively \textit{destroying} gaze-related features. 
{We believe this happens due to the \textbf{imbalanced sensitivity of $\mathcal{G}$ across different regions of the input image}. Specifically, $\mathcal{G}$ is significantly more sensitive to changes in the eye regions compared to other regions, as eye regions contain the most crucial features for gaze estimation~\cite{zhang17_cvprw}. As a result, perturbations added to these sensitive regions are more easily to cause substantial changes in the gaze estimation output. This imbalance causes the algorithm to prioritize adding perturbations to the sensitive eye regions when solving the optimization problem in Equation~\ref{opt:REforDRM}, neglecting potential trigger patterns in less sensitive regions. 


We address this issue by preventing significant changes in the gaze estimation output caused by perturbations added in sensitive regions in each training iteration, such that the algorithm can search for trigger patterns in both sensitive and insensitive regions. {Given an image $x_i$, we first estimate the sensitivity of $\mathcal{G}$ to each pixel in $x_i$ by computing the gradient of $\mathcal{G}$ with respect to that pixel. The intuition is that if $\mathcal{G}$ is sensitive to a pixel, e.g., pixels in the eye regions, a small change in its value will result in a significant change in the output of $\mathcal{G}$, which is reflected by a large absolute gradient value. By contrast, if $\mathcal{G}$ is insensitive to a pixel, the corresponding absolute gradient will be small. Formally, consider an image $x_i$ with dimensions $N_w \times N_h \times N_c$. We denote $x_i[a,b]$ as the pixel of $x_i$ at width $a$ and height $b$. The sensitivity $\mathcal{T}(x_i)[a,b]$ of this pixel is estimated as $\mathcal{T}(x_i)[a,b]=\sum_{c=1}^{N_c}{|{\partial \mathcal{G}}/{\partial x_i[a,b,c]}|}$, where $x_i[a,b,c]$ is the value of $x_i[a,b]$ in channel $c$. By computing the sensitivity for each pixel, we obtain a sensitivity map $\mathcal{T}(x_i)$ of size $N_w \times N_h$ for $x_i$.} We re-scale the sensitivity map to $[0,1)$ by dividing each component by a value greater than the maximum value in the map. Then, we obtained the reverse-engineered poisoned image $x'_i$ by:
\begin{equation}
\begin{split}
    x'_i[a,b,c]=M_{\theta}(x_i)[a,b,c]\cdot(1-\mathcal{T}(x_i)[a,b])+\\x_i[a,b,c]\cdot\mathcal{T}(x_i)[a,b],
    \label{eq:reversePoisoned}
\end{split}
\end{equation}{where $x'_i[a,b,c]$ and $M_{\theta}(x_i)[a,b,c]$ refer to the pixel value of $x'_i[a,b]$ and $M_{\theta}(x_i)[a,b]$ at channel $c$, respectively. Essentially, if $x_i[a,b]$ is sensitive, indicated by a large value of $\mathcal{T}(x_i)[a,b]$, we limit the perturbations added to it in each iteration.} Instead of directly feeding $M_{\theta}(x_i)$ to $\mathcal{G}$, we feed the image $ x'_i$ to $\mathcal{G}$ to form the final optimization problem $\mathcal{OPT}$-{\name} as: 
\begin{equation}
    \theta^{*}=\arg\min_{\theta} \frac{\lambda_1}{d}\sum_{j=1}^d{\sigma^2\left(\{\mathcal{G}_j\left(x'_i \right)\}_{i=1}^N\right)} +\lambda_2 r_f + \sum_{i=1}^N\frac{\|x'_i -x_i\|_1}{N},
    \label{opt:SecureGaze}
\end{equation} In a nutshell, $\mathcal{OPT}$-{\name} substitutes $M_{\theta}(x_i)$ in all the objectives of Equation~\ref{opt:REforDRM} with $x'_i$.}

\subsubsection{Backdoor identification}
By solving the new optimization problem defined in Equation~\ref{opt:SecureGaze}, we can obtain the perturbation $\|x'_i-x_i\|_1$ required to transform input $x_i$ to generate the potential target gaze direction. We observe that the perturbation needed to alter $x_i$ to produce the target gaze direction in a backdoored gaze estimation model is significantly smaller than that required for a benign gaze estimation model. To illustrate, we train ten benign and ten backdoored gaze estimation models using BadNets on the MPIIFaceGaze dataset. Figure~\ref{fig:averagePerturbation} shows the average perturbation on the benign dataset obtained by solving $\mathcal{OPT}$-{\name} for each model. The results show that the average perturbations required for the backdoored models (P0 to P9) are considerably smaller than those for the benign models (B0 to B9). 

{Based on this observation, we determine whether a given gaze estimation model is backdoored by comparing the average perturbation obtained through reverse engineering on $\mathcal{D}_{be}$ with a threshold value $\epsilon \|\hat{x}\|_1$. Here, $\hat{x}$ is the input image with the maximum $L1$ norm in the benign dataset $\mathcal{D}_{be}$, and $\epsilon$ is a constant. The average perturbation is calculated as $\frac{1}{N_{be}}\sum_{x_i\in\mathcal{D}_{be}}{\|x'_i-x_i\|_1}$, where $N_{be}$ represents the number of images in $\mathcal{D}_{be}$. To determine the threshold value, we assume that the perturbations of benign models follow a normal distribution. We compute the mean $m_p$ and standard deviation $\sigma_p$ of average perturbations across ten benign models reported in Figure~\ref{fig:averagePerturbation}. We set the threshold value to be $m_p - 2\sigma_p$, meaning that models with perturbation values below this threshold have a greater than 95\% probability of being outliers, indicating a backdoored model. This corresponds to $\epsilon=0.03$.}


\begin{figure}[]
	\centering
	{\includegraphics[scale=0.37]{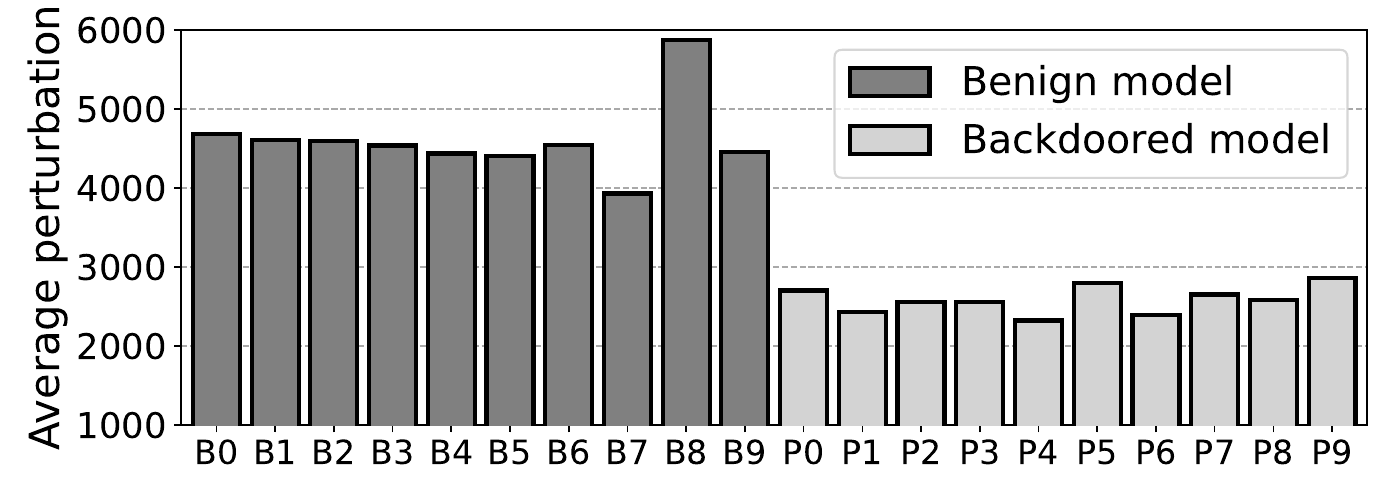}} 

\caption{The average perturbations for ten benign models (B0$\sim$B9) and ten backdoored models (P0$\sim$P9).}
	\label{fig:averagePerturbation}
 \vspace{-0.15in}
\end{figure}

\subsubsection{Backdoor mitigation} 
Once a gaze estimation model $\mathcal{G}$ is identified as backdoored, {\name fine-tunes $\mathcal{G}$ to mitigate backdoor behavior, such that the fine-tuned model produces correct gaze directions for poisoned images. \textbf{Note that, the defender only has access to a small benign dataset $\mathcal{D}_{be}$.} Therefore, \name generates a reverse-engineered poisoned dataset, $\mathcal{D}_{rp}=\{x'_i, y_i\}_{i=1}^{N_{be}}$, by applying $M_{\theta}$ to each image $x_i$ in $\mathcal{D}_{be}$ via Equation~\ref{eq:reversePoisoned}. Each reverse-engineered poisoned image $x'_i$ in $\mathcal{D}_{rp}$ is annotated with its correct gaze annotation $y_i$.} Next, {\name} fine-tunes $\mathcal{G}$ using both $\mathcal{D}_{be}$ and $\mathcal{D}_{rp}$. Formally, the backdoor mitigation is achieved by minimizing the following objective: $
    \sum_{(x_i,g_i)\in\mathcal{D}_{be}}\ell_1(\mathcal{G}(x_i),y_i) + \sum_{(x'_i,g_i)\in\mathcal{D}_{rp}}\ell_1(\mathcal{G}(x'_i),y_i)$.


\section{Evaluation}
\label{sec:evaluation}

\subsection{Evaluation Setups}

\subsubsection{Datasets} We consider two benchmark gaze estimation datasets that are collected in real-world settings. 

\begin{itemize}[leftmargin=*, wide, labelwidth=!, labelindent=0pt]
\setlength\itemsep{0.1em}
\item\textbf{MPIIFaceGaze}~\cite{zhang19_pami} is collected from 15 subjects during their routine laptop usage. Each subject contains 3,000 images under different backgrounds, illumination conditions, and head poses. 

\item\textbf{GazeCapture}~\cite{Krafka_2016_CVPR} is a large-scale dataset collected from over 1450 individuals in real-world environments. It comprises 2.5 million images captured using the front-facing cameras of smartphones, showcasing a diverse range of lighting conditions and backgrounds. 
\end{itemize}

{For each dataset, we randomly sample $80\%$ of the images to form the training dataset $\mathcal{D}_{tr}$ and $10\%$ to form the benign dataset $\mathcal{D}_{be}$, ensuring that there is no overlap between them.} $\mathcal{D}_{tr}$ is employed to train backdoored and benign models, while $\mathcal{D}_{be}$ is utilized for backdoor identification and mitigation. The remaining images constitute the testing set $\mathcal{D}_{te}$ to evaluate mitigation performance.

\subsubsection{Backdoor attacks} 

We consider {five} SOTA attacks, including both input-independent and input-aware attacks.

\begin{itemize}[leftmargin=*, wide, labelwidth=!, labelindent=0pt]
\setlength\itemsep{0.1em}
\item\textbf{BadNets}~\citep{badnet} generates poisoned inputs by pasting a fixed pattern as the backdoor trigger on the inputs. We use a 20 $\times$ 20 red patch located at the right-bottom corner as the backdoor trigger. 

\item\textbf{Clean Label}~\citep{turner2019label} exclusively applies a fixed pattern as the backdoor trigger to images belonging to the target class in classification. 
To adapt it for gaze estimation, we apply the trigger to images with gaze annotations ``close'' to $y_T$, specifically those where $|y - y_T| \leq \delta$. 

\item\textbf{WaNet}~\citep{nguyen2021wanet} generates stealth and input-aware backdoor triggers through image warping techniques. These triggers are injected into images using the elastic warping operation. 

\item\textbf{Input-aware dynamic attack (IADA)}~\citep{nguyen2020input} generates dynamic backdoor triggers using a trainable trigger generator, which produces backdoor triggers varying from input to input. 

\item{\textbf{Invisible backdoor attack (IBA)}~\citep{Li_2021_ICCV_ISSA} generates sample-specific invisible noises via steganography technique~\cite{Tancik_2020_CVPR} as the backdoor triggers, which contain information of a predefined string. 
}

\end{itemize}

\subsubsection{{Discussion on backdoor triggers}} {The backdoor triggers used in BadNets and Clean Label are input-independent, meaning their patterns and positions remain fixed across different inputs. In our setup, we consider a red patch in the bottom-right corner as the backdoor trigger. However, an attacker could use various patterns in different locations. As long as the pattern and position of the trigger remain consistent during both training and inference, the attacker can successfully execute a backdoor attack.}

{In contrast, WaNet, IADA, and IBA employ input-aware triggers, where the patterns and positions vary across different inputs. Figure~\ref{fig:iada_poiosned} illustrates poisoned images generated by IADA, showcasing how the triggers change from one input to another.
Additionally, WaNet and IBA create imperceptible backdoor triggers using image warping and DNN-based steganography techniques, respectively. To demonstrate the invisibility of these triggers, Figure~\ref{fig:invisible_trigger} presents both benign and poisoned images produced by WaNet and IBA.}

\begin{figure}[]
\begin{minipage}[c]{\linewidth}
	\centering
	\includegraphics[scale=0.585]{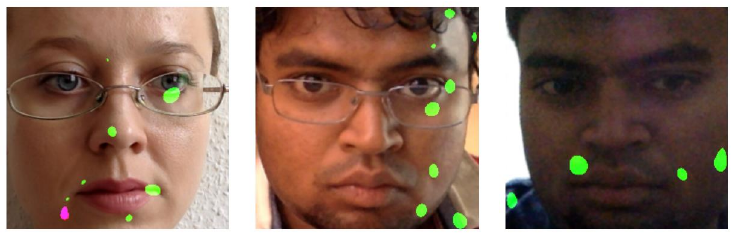} 
        \caption{{Poisoned images generated by IADA. The patterns and positions of triggers vary across different inputs.}}	
	\label{fig:iada_poiosned}
\end{minipage}

\begin{minipage}[c]{\linewidth}
	\centering
	\includegraphics[scale=0.365]{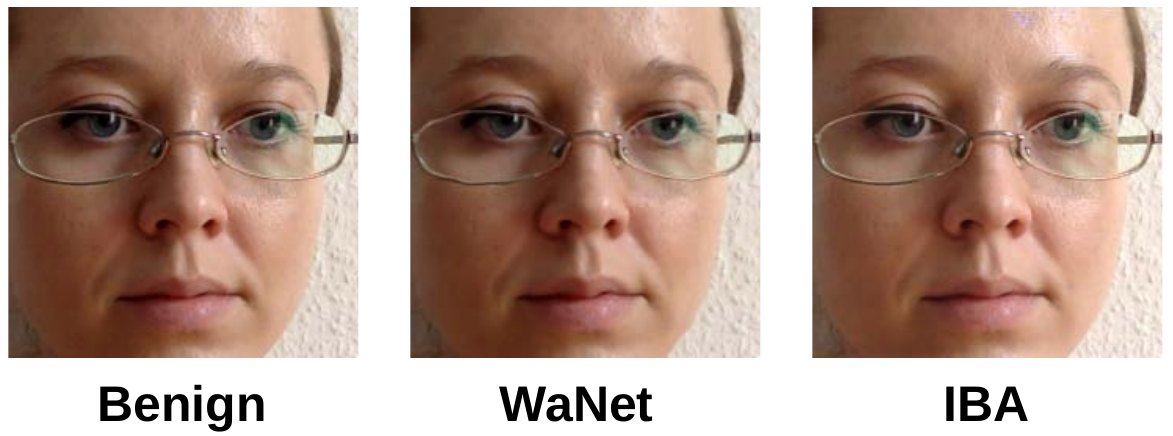} 
        \caption{{Comparsion between benign image and images poisoned by WaNet and IBA. The triggers are invisible.}}	
	\label{fig:invisible_trigger}
    \end{minipage}
\end{figure}

\subsubsection{{Attack time overhead}} {Below, we analyze the time overhead required to launch these attacks. For attacks in the digital world, the attacker needs to inject triggers into images. To quantify this, we measure the latency of trigger injection for each attack on a desktop equipped with an NVIDIA GeForce RTX 3080 Ti GPU and an Intel i7-12700KF CPU, with results presented in the Table below:

\begin{table}[h]
\resizebox{\linewidth}{!}{%
\begin{tabular}{c|ccccc}
{Attack}  & {BadNets} & {Clean Label} & {IADA}    & {WaNet} & {IBA} \\ \Xhline{2\arrayrulewidth}
{Latency} & {0.3 ms}  & {0.3 ms}     & {12.3 ms}&  {4.5 ms}      & {1.9 ms}    
\end{tabular}%
}
\vspace{-0.1in}
\label{Tab:attack_latency}
\end{table}

As shown, BadNets requires an overhead of 0.3 ms for applying a fixed pattern directly to the image. In contrast, IADA incurs a significantly higher overhead of 12 ms due to the use of generative models for trigger injection. For attacks in the physical world (such as the one we demonstrated in Section~\ref{subsubsec:physical_attack}), the time required to place the physical trigger within the camera view is negligible. For both digital and physical world attacks, once the trigger appears in the camera view or image, the backdoored model exhibits its backdoored behavior with the same latency as a standard model inference, e.g., 12 ms for a model implemented using ResNet18~\citep{He_2016_CVPR}.}

\subsubsection{Compared defenses}
We compare our method with the following defenses: 

\begin{itemize}[leftmargin=*, wide, labelwidth=!, labelindent=0pt]
\setlength\itemsep{0.3em}

\item\textbf{Gaze-NC} is 
adapted from NC~\citep{wang2019neural}. 
Since NC is designed for classification and needs to enumerate all potential targets, we adapt it for gaze estimation by treating the potential target gaze direction as the optimization variable.

\item\textbf{Gaze-FRE} is adapted from FRE~\citep{wang2022rethinking}, which utilizes the feature-space characteristics of backdoored classification models. {Similar to Gaze-NC}, we adapt it for gaze estimation by considering the potential target gaze direction as an optimization variable.

\item\textbf{Fine-prune}~\citep{liu2018fine} notes that the compromised neurons for backdoored classification models are dormant for benign inputs. Therefore, given a benign dataset, Fine-prune removes neurons with low activation values for benign images. 

\item\textbf{ANP}~\citep{wu2021anp} observes that the compromised neurons are sensitive to perturbations. 
Based on this observation, ANP applies adversarial attacks to the neurons to identify sensitive neurons and subsequently prunes them for backdoor defense.

\item{\textbf{NAD}~\citep{li2021neural} first fine-tunes the given backdoored model on the benign dataset. It then treats the fine-tuned model as a teacher model and performs knowledge distillation to the original model.}

\item {\textbf{RNP}~\citep{RNP} first maximizes errors on clean samples at the neuron level, then minimizes errors on the same samples at the filter level to identify compromised neurons for pruning.}

\item\textbf{Fine-tune} serves as a straightforward baseline that employs the benign dataset to directly fine-tune the backdoored models. We consider this baseline as existing research~\cite{liu2018fine, wu2021anp} show its effectiveness on backdoor mitigation.

\end{itemize}

\subsubsection{Evaluation metrics}
Given a set of benign and backdoored 
models, we use the following metrics to evaluate the performance of \name on backdoor identification:
\begin{itemize}[leftmargin=*, wide, labelwidth=!, labelindent=0pt]
\setlength\itemsep{0.1em}
    \item \textbf{Identification Accuracy (Acc):} the percentage of correctly classified models (either benign or backdoored) over all the models.
    \item \textbf{True Positives (TP):} the number of correctly identified backdoored gaze estimation models.
    \item \textbf{False Positives (FP):} the number of benign gaze estimation models recognized as backdoored models.
    \item \textbf{False Negatives (FN):} the number of backdoored gaze estimation models identified as benign models.
    \item \textbf{True Negatives (TN):} the number of correctly recognized benign gaze estimation models. 
    \item \textbf{ROC-AUC:} the ROC-AUC score computed from the average perturbations 
    for benign and backdoored gaze estimation models. This metric is used to compare the backdoor identification performance between {\name}, Gaze-NC, and Gaze-FRE.
\end{itemize}

\noindent To evaluate the performance on backdoor mitigation, we generate a poisoned dataset $\mathcal{PD}_{te}$ by applying the trigger function to all the images in $\mathcal{D}_{te}$. Then, we use the following metrics:
\begin{itemize}[leftmargin=*, wide, labelwidth=!, labelindent=0pt]
\setlength\itemsep{0.1em}
    \item \textbf{Average Attack Error (AE):} the average angular error between the estimated gaze directions and the target gaze directions over all the images in $\mathcal{PD}_{te}$. 
    \item \textbf{Defending Attack error (DAE):} the average angular error between the estimated gaze directions and the correct gaze annotations over all the images in $\mathcal{PD}_{te}$.
\end{itemize}
A larger AE and a smaller DAE indicate better mitigation performance, while a smaller AE indicates better attack performance. 



\begin{table}[]\huge
\caption{\textbf{Backdoor identification performance on MPIIFaceGaze and GazeCapture for different attacks}. {\name} can identify the backdoored gaze estimation models on two datasets with over 92\% accuracy.
}
\resizebox{\linewidth}{!}{
\begin{tabular}{cccccclccccc}
\Xhline{3\arrayrulewidth}\\[-2.5ex]
\multirow{2}{*}{Attack} & \multicolumn{5}{c}{\textbf{MPIIFaceGaze}} &  & \multicolumn{5}{c}{\textbf{GazeCapture}} \\ \cline{2-6} \cline{8-12} \\[-2.5ex]
                        & TP    & FP    & FN    & TN    & Acc       &  & TP    & FP    & FN    & TN    & Acc      \\ \hline\\[-2.5ex]
BadNets                 & 20    & 3     & 0     & 17    & 92.5\%    &  & 20    & 2     & 0     & 18    & 95.0\%   \\
IADA                    & 20    & 3     & 0     & 17    & 92.5\%    &  & 19    & 2     & 1     & 18    & 92.5\%   \\
Clean Label             & 20    & 3     & 0     & 17    & 92.5\%    &  & 20    & 2     & 0     & 18    & 95.0\%   \\
WaNet                   & 20    & 3     & 0     & 17    & 92.5\%    &  & 20    & 2     & 0     & 18    & 95.0\%   \\
IBA                     & 20    & 3     & 0     & 17    & 92.5\%    &  & 20    & 2     & 0     & 18    & 95.0\%   \\ \Xhline{3\arrayrulewidth}
\end{tabular}}
\label{Tab_evaluation_results_MPII}
\vspace{-0.1in}
\end{table}


\begin{table*}[]
\centering
\caption{\textbf{The ROC-AUC scores of different backdoor identification methods when evaluating on MPIIFaceGaze and GazeCapture with different attacks}. {\name} outperforms both Gaze-NC and Gaze-FRE significantly.
}
\resizebox{0.89\linewidth}{!}{
\begin{tabular}{cccccccccccc}
\Xhline{2\arrayrulewidth}\\[-2.5ex]
\multirow{2}{*}{Method} & \multicolumn{5}{c}{MPIIFaceGaze}              &  & \multicolumn{5}{c}{GazeCapture}               \\ \cline{2-6} \cline{8-12} \\[-2.5ex]
                        & BadNets & IADA  & Clean Label & WaNet & All   &  & BadNets & IADA  & Clean Label & WaNet & All   \\ \hline\\[-2.5ex]
Gaze-NC                 & 0.400   & 0.311 & 0.002       & 0.828 & 0.385 &  & 0.417   & 0.605 & 0.026       & 0.630 & 0.419 \\
Gaze-FRE                & 0.561   & 0.512 & 0.444       & 0.508 & 0.506 &  & 0.528   & 0.531 & 0.461       & 0.601 & 0.530 \\
SecureGaze              & \textbf{0.995}   & \textbf{1.000} & \textbf{1.000}       & \textbf{0.995} & \textbf{0.998} &  & \textbf{0.995}   & \textbf{1.000} & \textbf{1.000}       & \textbf{0.967} & \textbf{0.986} \\ \Xhline{2\arrayrulewidth}
\end{tabular}}
\label{Tab_AUC_MPII}
\vspace{-0.05in}
\end{table*}

\begin{table*}[]\large
\caption{\textbf{Performance of \name in backdoor mitigation}. 
\name shows larger AE and smaller DAE on two datasets, which demonstrates good performance in backdoor mitigation for various attacks.
}

\label{Tab:backdoor_mitigation_diff_attack}
\resizebox{0.88\textwidth}{!}{%
\begin{tabular}{ccccccccccccc}
\Xhline{2\arrayrulewidth}\\[-2.5ex]
\multirow{2}{*}{Method}     & \multirow{2}{*}{Metric} & \multicolumn{5}{c}{MPIIFaceGaze}     &  & \multicolumn{5}{c}{GazeCapture}      \\ \cline{3-7} \cline{9-13} \\[-2.5ex]
                            &                         & BadNets & IADA & Clean Label & WaNet & {IBA} &  & BadNets & IADA & Clean Label & WaNet & {IBA}\\ \hline\\[-2.5ex]
\multirow{2}{*}{Undefended} & AE                      & 3.25    & 3.19 & 0.72        & 1.31  & {3.04} &  & 1.09    & 1.54 & 2.45        & 2.51  & {0.91}\\
                            & DAE                     & 14.8    & 14.4 & 15.4        & 15.9  & {14.4} &  & 20.0    & 10.6 & 9.85        & 9.55  & {19.4}\\ \hline\\[-2.5ex]
\multirow{2}{*}{{\name}}    & AE                      & 17.2    & 15.6 & 16.4        & 15.3  & {14.2} &  & 17.7    & 10.2 & 10.9        & 9.57  & {19.2}\\
                            & DAE                     & 3.59    & 3.50 & 2.51        & 3.29  & {4.12} &  & 3.65    & 3.77 & 3.20        & 3.66  & {3.90}\\ \Xhline{2\arrayrulewidth}
\end{tabular}%
}
\vspace{-0.05in}
\end{table*}

\subsubsection{Implementation}
\label{sec:implementation}
We develop \name using TensorFlow and Adam optimizer~\cite{kingma2014adam}. We use a simple auto-encoder to implement $M_{\theta}$, which is similar to that used in~\citep{nguyen2020input}. Before performing the reverse engineering, we pre-train $M_{\theta}$ on the benign dataset $\mathcal{D}_{be}$ for 5,000 steps with the learning rate of 0.001. We train $M_{\theta}$ for 2,000 steps with a batch size of 50 and the learning rate of 0.0015. 
We set $\lambda_1=20$, $\lambda_2=800$. 
For backdoor mitigation, we fine-tune the gaze estimation models using a batch of 50 benign and 50 reverse-engineered poisoned images for 300 iterations.
We use ResNet18 \citep{He_2016_CVPR} (without the dense layer) to implement $F$, and a dense layer without activation function to implement $H$.

\subsection{Backdoor Identification Performance}
{We evaluate backdoor identification performance on 200 backdoored and 40 benign gaze estimation models. Specifically, for each dataset, we first train 20 benign models and then train 20 backdoored models for each attack. It is important to note that although the 20 backdoored (or benign) models for each attack-dataset combination are trained on the same training dataset, they have different parameters and exhibit variations in performance due to two key factors: 1) Each model is randomly initialized with different parameters; 2) During training, image batches are randomly sampled in each iteration, introducing variability in the training process. This is a standard evaluation protocol used in existing works~\cite{wang2022rethinking,Feng_2023_CVPR}.}

\vspace{2pt}
\noindent\textbf{Evaluation results.} 
We report the backdoor identification results of \name in Table \ref{Tab_evaluation_results_MPII}, which indicate that {\name} can identify backdoored gaze estimation models trained by both input-independent and input-aware attacks, on MPIIFaceGaze and GazeCapture, with an average accuracy of 92.5\% and 94.5\%, respectively. Specifically, TP and FN remain consistent across most evaluation scenarios, with TP being 20 and FN being 0. This indicates that \name successfully identifies all 20 backdoored gaze estimation models without any false negatives. Moreover, TN and FP are identical for each backdoor attack. This consistency arises because the set of benign gaze estimation models, which are attack-free and independent of backdoor attack, remains the same across all five scenarios, and thus, \name leads to the same identification results, in FP and TN, regardless of the attacks. 

\vspace{2pt}
\noindent\textbf{{Discussion on failure cases.}}
{For FN cases, \name struggles to identify a trigger function that produces similar gaze directions, instead prioritizing the minimization of perturbations. By contrast, for FP cases, \name reverse-engineers a trigger function that maps different inputs close to the target gaze direction but neglects the magnitude of perturbations introduced to benign images. We believe this issue arises from using fixed values for $\lambda_1$ and $\lambda_2$, where FN cases require larger values, while FP cases benefit from smaller values. 
A potential solution is to dynamically adjust $\lambda_1$ and $\lambda_2$. For instance, we can initially increase their values to ensure the trigger function generates similar outputs across different inputs, then gradually decrease them to focus on minimizing perturbations.}

\vspace{2pt}
\noindent\textbf{{Comparison with state-of-the-art defenses.}}
Table \ref{Tab_AUC_MPII} shows the ROC-AUC scores of {\name}, Gaze-NC, and Gaze-FRE for different backdoor attacks on two datasets. We also report the scores when applying the various attacks simultaneously. 
As shown, the score of {\name} is above $0.96$ in all the examined cases, which is significantly higher than that of Gaze-NC and Gaze-FRE. Besides, we notice that Gaze-FRE fails to find a trigger function that enables the backdoored gaze estimation model to map different inputs to similar gaze directions. This observation confirms our analysis that the feature-space characteristics for backdoored classification models~\citep{wang2022rethinking} do not hold for backdoored gaze estimation models.

\subsection{Backdoor Mitigation Performance}


\noindent\textbf{Evaluation results.} We train backdoored gaze estimation models by each considered attack on each dataset. The backdoor mitigation results of \name are shown in Table \ref{Tab:backdoor_mitigation_diff_attack}, which indicate that {\name} can mitigate backdoor behaviors for various attacks on two dataset. Specifically, {\name} can substantially increase AE, indicating that the output gaze directions for poisoned inputs deviate significantly from the target gaze direction after backdoor mitigation. Additionally, {\name} significantly reduces DAE, which means that the mitigated backdoored gaze estimation models perform normally and output gaze directions close to correct gaze annotation, even though triggers are injected into the inputs.

\vspace{2pt}
\noindent\textbf{{Comparison with state-of-the-art defenses.}} We compare {\name} with ANP, RNP, NAD, Fine-prune, and Fine-tune on backdoor mitigation. Specifically, we train a backdoored gaze estimation model by WaNet on MPIIFaceGaze and apply different methods to mitigate the backdoor behavior. The evaluation results are shown in Figure \ref{fig:mitigation_comparsion}. The AE for {\name} is significantly larger than that for other methods, while the DAE for {\name} is much smaller than that for other methods, which shows the superiority of {\name} on backdoor mitigation. Moreover, Fine-prune, ANP, and RNP, which prune compromised neurons, perform poorly on backdoored gaze estimation models. This supports our analysis that the feature-space characteristics of backdoored gaze estimation models differ from those of backdoored classification models, making it ineffective to target specific neurons for backdoor mitigation. 

\begin{figure}[]
	\centering
    \includegraphics[scale=0.2]{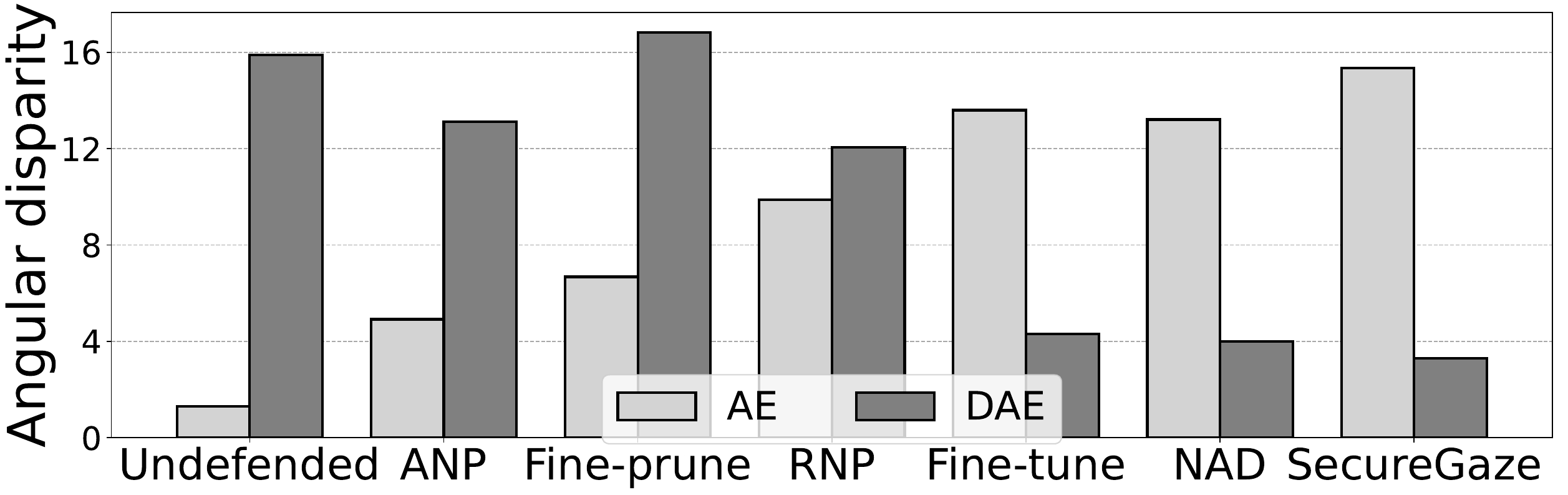}
    \caption{\textbf{Performance of different defenses on backdoor mitigation.} {\name} outperforms compared methods.}
    \vspace{-0.15in}
    \label{fig:mitigation_comparsion}
\end{figure}

\begin{table*}[]

\begin{minipage}[c]{0.63\linewidth}\Large

\centering
\caption{The impact of FSO, $\lambda_1$, $\lambda_2$, and $p$ on backdoor identification performance. }
\resizebox{\linewidth}{!}{%
\begin{tabular}{ccccccccccccc}
\Xhline{3\arrayrulewidth}\\[-2ex]
\multirow{2}{*}{Metric} & \multicolumn{3}{c}{Different $\lambda_1$} &  & \multicolumn{3}{c}{ Different $\lambda_2$} & & \multicolumn{3}{c}{Different $p$} & \multirow{2}{*}{w/o FSO}\\ \cline{2-4} \cline{6-8} \cline{10-12} \\[-2ex]
                        & 10      & 20     & 30     &  & 600     & 800    & 1000  &  & $5\%$     & $10\%$    & $15\%$  \\ \hline\\[-2ex]
TP                      & 20      & 20     & 19     &  & 20      & 20     & 20    &  & 20        & 20        & 20     & 20  \\
FP                      & 3       & 3      & 3      &  & 11      & 3      & 3     &  & 4         & 3         & 3      & 20 \\
FN                      & 0       & 0      & 1      &  & 0       & 0      & 0     &  & 0         & 0         & 0      & 0 \\
TN                      & 17      & 17     & 17     &  & 9       & 17     & 17    &  & 16        & 17        & 17     & 0  \\
Acc                     & 92.5\%  & 92.5\% & 90\%   &  & 72.5\%  & 92.5\% & 92.5\%&  & 90\%      & 92.5\%    & 92.5\% & 50\% \\ \Xhline{3\arrayrulewidth}
\end{tabular}
\label{Tab:abliation_study}%
\vspace{-0.1in}
}
\end{minipage}
\quad
%
\begin{minipage}[c]{.33\linewidth}\footnotesize
\centering
\caption{Results on adaptive attack with different values for $\beta$. Adaptive WaNet is less effective than WaNet. }
\label{Tab:adaptive_attack}
\resizebox{0.94\textwidth}{!}{%
\begin{tabular}{cccc}
\Xhline{2\arrayrulewidth}\\[-2.5ex]
\multirow{2}{*}{Metric} & \multirow{2}{*}{WaNet} & \multicolumn{2}{c}{Adaptive WaNet} \\ \cline{3-4} \\[-2ex]
                        &                        & $\beta=$0.02                 &  $\beta=$1.0         \\ \hline
AE                      &  1.50                  & 5.41                 &  10.8         \\
DAE                     &  16.0                  & 14.9                 &  12.8         \\
Acc                     &  92.5\%                & 92.5\%               &  67.5\%     \\\Xhline{2\arrayrulewidth}
\end{tabular}%
}
\end{minipage}
\vspace{-0.1in}
\end{table*}

\subsection{{System Profiling}}

{We measure the latency and memory usage of \name during two key processes: reverse-engineering the trigger function for backdoor identification and fine-tuning the model for backdoor mitigation. These measurements are conducted on a desktop equipped with an NVIDIA GeForce RTX 3080 Ti GPU and an Intel i7-12700KF CPU, following the implementation details outlined in Section~\ref{sec:implementation}.} 

\vspace{2pt}
\noindent{\textbf{Latancy. }For reverse-engineering the trigger function, we repeat the process five times for a given gaze estimation model. The average latency for reverse engineering is 12 minutes. For backdoor mitigation, we repeat the experiments five times with a given gaze estimation model and a reverse-engineered trigger function. The average latency for backdoor mitigation is 100 seconds.}

\vspace{2pt}
\noindent{\textbf{Memory usage.} We measure the memory specifically allocated to the training process, i.e., training $M_{\theta}$ or fine-tuning the gaze estimation model. This is determined by subtracting the memory usage before training from the peak memory usage during training. Specifically, reverse-engineering the trigger function requires approximately 9,970 MB of memory, while fine-tuning the gaze estimation model consumes around 6,000 MB.}

\vspace{2pt}
Note that, for a given gaze estimation model, the process of backdoor identification and mitigation needs to be performed offline only once before deployment. As a result, \name does not introduce additional run-time latency to the gaze estimation model after deployment. Furthermore, since \name does not need to enumerate all potential targets, it is more efficient than existing reverse-engineering-based techniques that require scanning all labels (e.g., 140 minutes for FRE~\citep{wang2022rethinking} on ImageNet~\citep{deng2009imagenet}).

\subsection{Ablation Studies}

We conduct ablation studies to investigate the impact of different design choices on the performance of \name. We consider WaNet as the attack and evaluate on MPIIFaceGaze.

\subsubsection{Impact of weights and the size of benign dataset}
We vary the values of $\lambda_1$ and $\lambda_2$ in Equation \ref{opt:REforDRM} from $10$ to $30$ and from $600$ to $800$ respectively to investigate their impacts on the performance of backdoor identification. Moreover, 
we study the impact of the size of $\mathcal{D}_{be}$ on the identification performance by changing the ratio $p$ of the benign dataset to the original whole dataset from $5\%$ to $15\%$. 
We report the backdoor identification results in Table~\ref{Tab:abliation_study}. We observe that the performance of {\name} is insensitive to $\lambda_1$, as the identification accuracy is almost stable with different $\lambda_1$. However, {\name} is sensitive to $\lambda_2$ and the identification accuracy increases with $\lambda_2$, as a larger $\lambda_2$ allows the feature-space optimization objective to have a greater contribution to the optimization problem.
This observation proves that the proposed feature-space optimization objective is important for backdoor identification. 
Additionally, as $p$ decreases from $10\%$ to $5\%$, the identification accuracy and TN decrease, while TP remains stable.
This is because, compared to a larger $p$, it is easier to find a small amount of perturbation that can lead to the backdoor behavior on a smaller $p$ for benign models. However, the identification accuracy is still $90\%$ even when $p=5\%$.

\subsubsection{Impact of feature-space optimization objective (FSO)} 
We study the impact of FSO on the performance of backdoor identification by removing it from $\mathcal{OPT}$-{\name}. The results are shown in Table \ref{Tab:abliation_study}, which indicates that all the backdoored and benign models are classified as backdoored models. This means that {\name} cannot identify backdoor without the FSO. 
We further observe that without the FSO, {\name} cannot find a trigger function that can map different inputs to similar output vectors. As a result, {\name} solves the optimization problem by focusing on minimizing the amount of perturbations, which leads to the misclassification of backdoored models.

\subsubsection{{Generalization to various datasets.}}
{We investigate the generalization capability of \name across different regression tasks by utilizing three additional datasets. These include XGaze~\citep{Zhang2020ETHXGaze}, a complex gaze estimation dataset with a wider range of head poses and gaze directions, and two datasets focused on head pose estimation, i.e., Biwi~\cite{Biwi} and Pandora~\cite{Pandora}. Specifically, 
head pose estimation seeks to determine a three-dimensional vector representing the Eular angle (yaw, pitch, roll) from a monocular image. This modality is commonly used in human-computer interaction to infer user attention~\cite{hueber2020headbang,crossan2009head,wang2022faceori} and for authentication system~\cite{liang2021auth}. Biwi is collected from 24 subjects, and each subject has 400 to 900 images. We use the \emph{cropped faces of Biwi dataset (RGB images)} released by \cite{Pandora}. Pandora has 100 subjects and more than 120,000 images.
}

{
We train 20 backdoored and 20 benign models on the training dataset for each dataset. For head pose estimation, we set $\lambda_1=10$, $\lambda_2=100$, and $\epsilon=0.05$, using average $L_1$ error to define AE and DAE, rather than average angular error. The evaluation results for backdoor identification and mitigation are shown in Table~\ref{Tab_different_Dataset}, which demonstrates that \name is effective across various datasets and regression tasks in human-computer interaction.}




\subsection{Adaptive Attack}

When the attacker has the full knowledge of {\name}, one potential adaptive attack that can bypass our method is to force RAV to be close to 1 to break the feature-space characteristics. Based on this intuition, we design an adaptive attack that adds an additional loss term $L_{adp}$ with a weight $\beta$ to the original loss function of the chosen attack to enforce RAV to be close to one. We define $L_{adp}$ as:
\begin{equation}
    L_{adp}= \Big|1-\frac{1}{d}\sum_{j=1}^{d}{\frac{{
    \sigma^2\left(\big\{\mathcal{B}(F(\mathcal{A}(x_i)), w_j)\big\}_{i=1}^{N_p} \right)}}{{\sigma^2\left(\big\{\mathcal{B}(F((x_i), w_j))\big\}_{i=1}^{N_b} \right)}}}\Big|,
\end{equation}where $N_p$ and $N_b$ are the numbers of poisoned and benign inputs in a minibatch. 
We consider two values for $\beta$, i.e., 0.02 and 1.0. We train 20 backdoored models by incorporating $L_{adp}$ into WaNet for each considered value of $\beta$.

Table \ref{Tab:adaptive_attack} shows the identification accuracy and the averaged AE and DAE over 20 backdoored models.  
As shown, the AE of the adaptive attack is much higher than that of WaNet {and it increases as $\beta$ raises.} This proves that the feature-space characteristics we observed of backdoored gaze estimation models are vital to result in the backdoor behavior. {Moreover, the adaptive attack with $\beta=0.02$ cannot reduce the identification accuracy as the feature-space characteristics are not totally broken. When increasing $\beta$ to 1.0, the identification accuracy drops to 67.5\%. However, the AE is higher than $10.0$ with $\beta=1.0$, in which we believe the attack is ineffective.}

\begin{table}[]\Large
\caption{{Backdoor identification and mitigation performance of {\name} on 
various datasets. \name is effective across various datasets and regression tasks.}}
\label{Tab_different_Dataset}
\resizebox{\linewidth}{!}{%
\begin{tabular}{cccccc|cccc}
\multicolumn{1}{c|}{\textbf{}}                                          & \multicolumn{5}{c|}{\textbf{{Backdoor identification}}}                                                                 & \multicolumn{4}{c}{\textbf{{Backdoor mitigation}}}                         \\ \\[-2.5ex]\Xhline{3\arrayrulewidth}\\[-2.5ex]
\multicolumn{1}{c|}{\multirow{2}{*}{{Dataset}}} & \multirow{2}{*}{{TP}} & \multirow{2}{*}{{FP}} & \multirow{2}{*}{{FN}} & \multirow{2}{*}{{TN}} & \multirow{2}{*}{{Acc}} & \multicolumn{2}{c}{{Undefended}} & \multicolumn{2}{c}{{SecureGaze}} \\ 
\multicolumn{1}{c|}{}                         &                     &                     &                     &                     &                      & {AE}             & {DAE}           & {AE}             & {DAE}           \\ \\[-2.5ex]\Xhline{3\arrayrulewidth}\\[-2.5ex]
\multicolumn{1}{c|}{{XGaze}}                & {20}                  & {0}                   & {0}                   & {20}                  & {100\%}                  & {1.24}           & {44.5}          & {45.4}           & {2.97}          \\
\multicolumn{1}{c|}{{Biwi}}              & {20}                  & {0}                   & {0}                   & {20}                  & {100\%}                  & {2.48}           & {25.2}          & {27.1}           & {1.10}          \\
\multicolumn{1}{c|}{{Pandora}}                  & {20}                  & {0}                   & {0}                   & {20}                  & {100\%}                 & {0.48}           & {22.8}          & {23.0}           & {2.71}          \\ 
\end{tabular}%
}
\vspace{-0.04in}
\end{table}

\subsection{Physical-world Backdoor Defense}

Below, we apply \name to mitigate the backdoor behavior of the backdoored gaze estimation model we considered in Section~\ref{subsubsec:physical_attack}, in which a physical item, i.e., a piece of white tape on the face, can effectively trigger the backdoored behaviors. We record the estimated gaze from the mitigated model for each subject by the gaze estimation pipeline in Figure~\ref{fig:gaze_pipeline} when the physical trigger is present on the face. Figure~\ref{fig:mitigated_poisoned_gaze} visualizes the estimated gaze directions, {while Table~\ref{tab:physical_attack_mitigation} quantifies the average attack error, comparing results before and after mitigation.} Specifically, before mitigation, the estimated gaze directions (green dots) are concentrated around the attacker-chosen target (presented by the red star), {exhibiting a small average attack error}. By contrast, after backdoor mitigation using {\name}, the gaze estimations (blue dots) form four clusters corresponding to the four corners where the stimulus appeared, {resulting in a significantly higher average attack error than before migitation}. Moreover, we also plot the gaze directions estimated by the backdoored model from subjects without triggers (yellow dots) in Figure~\ref{fig:mitigated_poisoned_gaze}, which overlap with the estimations for subjects with physical triggers after mitigation (blue dots), indicating that \name can effectively mitigate the backdoor behavior. A video demo that compares behaviors of the backdoored and backdoor-mitigated models can be found in our GitHub repository: \url{https://github.com/LingyuDu/SecureGaze}.

\begin{figure}[]
    \centering
    \includegraphics[scale=0.53]{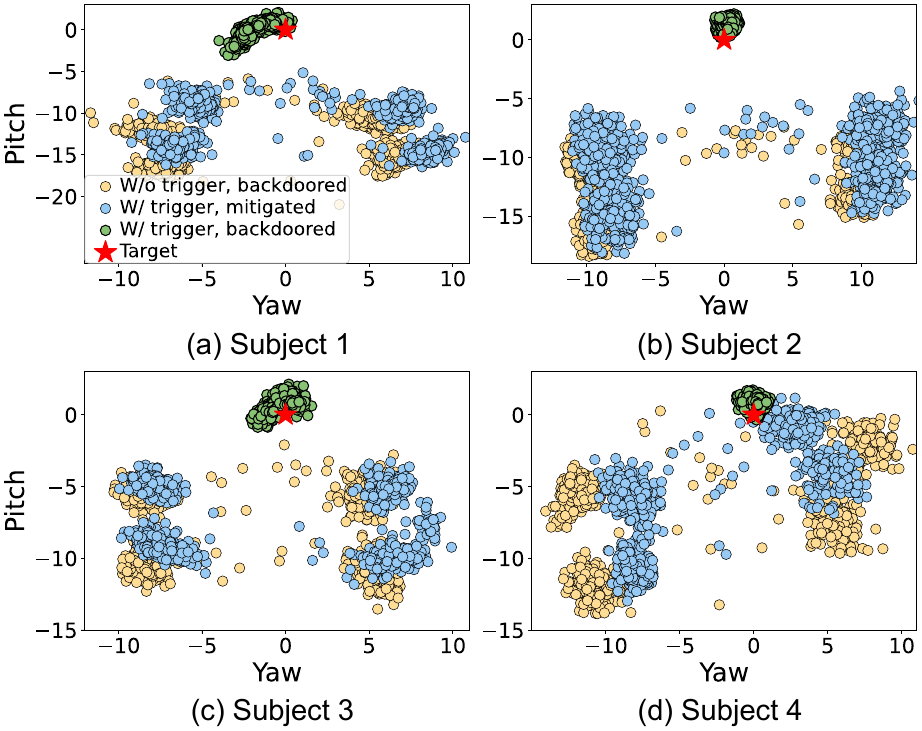}
    \caption{Gaze directions estimated by the backdoored gaze estimation model before (green dots) and after (blue dots) backdoor mitigation using \name.
    }
    \label{fig:mitigated_poisoned_gaze}
\end{figure}

\begin{table}[]
\caption{{The average attack error (in degree) for subjects with physical triggers before and after backdoor mitigation.}}
\resizebox{\linewidth}{!}{
\begin{tabular}{ccccc}
\Xhline{2\arrayrulewidth}
{Model}                & {Subject 1} & {Subject 2} & {Subject 3} & {Subject 4} \\ \hline
{Before mitigation}    & {1.71}      & {1.07}      & {0.98}      & {1.17}      \\
{After mitigation}     & {15.9}      & {16.6}      & {10.3}      & {7.6}      \\ \Xhline{2\arrayrulewidth}
\end{tabular}}
\label{tab:physical_attack_mitigation}
\vspace{-0.1in}
\end{table}


\subsection{{Limitations and Future Works}}

\noindent{\textbf{Limitation.} Similar to existing reverse-engineering-based methods~\citep{wang2022rethinking,Feng_2023_CVPR}, the current design of \name adopts a fixed threshold for backdoor identification, which may be less effective if the attacker employs a large trigger. 
Moreover, while we have investigated the impact of different hyperparameter values on the performance of \name, our analysis is limited to a few scenarios rather than an exhaustive search across a broader range of cases.}

\vspace{2pt}
\noindent{\textbf{Future research directions.} A promising avenue for future research involves extending \name to a wider range of regression models with continuous output spaces that are adopted in human-computer interactions.
Another interesting direction is to generalize \name to more complex threat scenarios, e.g., the gaze estimation models are backdoored by multiple trigger functions associated with multiple target gaze directions. Additionally, exploring more adaptive attacks could provide deeper insights into the robustness and limitations of \name, enabling a more comprehensive investigation into backdoor defenses tailored specifically for gaze estimation models.}

\section{Conclusion}
\label{sec:conclusion}
In this paper, we present \name, the first approach to defend gaze estimation models against backdoor attacks. We identify the unique characteristics of backdoored gaze estimation models, based on which we introduce a novel suite of techniques to reverse engineer the trigger function for backdoored gaze estimation models without the need to enumerate all the outputs. Our comprehensive experiments in both digital and physical worlds show that \name is consistently effective in defending gaze estimation models against six backdoor attacks that are triggered by input-aware patterns, input-independent patterns, and physical objects. We also adapt seven state-of-the-art classification defenses, showing that they are ineffective for gaze estimation, while {\name} consistently outperforms them. 

\balance
\bibliographystyle{ACM-Reference-Format}
\bibliography{09reference}


\end{document}